\documentclass[10pt,twocolumn,letterpaper]{article}

\usepackage[pagenumbers]{cvpr} 

\definecolor{cvprblue}{rgb}{0.21,0.49,0.74}
\usepackage[pagebackref,breaklinks,colorlinks,allcolors=cvprblue]{hyperref}

\usepackage[utf8]{inputenc} 
\usepackage[T1]{fontenc}    
\usepackage{url}            
\usepackage{booktabs}       
\usepackage{amsfonts}       
\usepackage{nicefrac}       
\usepackage{microtype}      
\usepackage{xcolor}         
\usepackage{graphicx}
\usepackage{wrapfig}
\usepackage{mathtools}
\usepackage{amsthm}
\usepackage{enumitem}
\usepackage{subcaption}
\usepackage{natbib}
\usepackage[capitalize,noabbrev]{cleveref}
\usepackage{algorithm}
\usepackage{algpseudocode}
\usepackage{multirow}
\usepackage{soul} 
\usepackage{bbm}
\usepackage{graphicx}
\usepackage{float}
\usepackage{colortbl}      
\definecolor{RoyalBlue}{rgb}{0.25, 0.41, 0.88}
\newcommand{\cellhi}{\cellcolor{RoyalBlue!15}}

\theoremstyle{plain}
\newtheorem{theorem}{Theorem}[section]

\newtheorem{corollary}[theorem]{Corollary}
\theoremstyle{definition}

\newtheorem{remark}[theorem]{Remark}

\definecolor{darkbrown}{RGB}{120,70,20}   
\definecolor{darkgreen}{RGB}{34,85,34}    

\def\paperID{21994} 
\def\confName{CVPR}
\def\confYear{2026}

\title{MFM-point: Multi-scale Flow Matching for Point Cloud Generation}

\author{%
  Petr Molodyk$^{1,}$\thanks{Equal contribution. Correspondence to: Yongxin Chen}\ \ , Jaemoo Choi$^{1,*}$, David W. Romero$^{2}$, Ming-Yu Liu$^{2}$, Yongxin Chen$^{1,2}$  \\[3pt]
  $^1$Georgia Institute of Technology, $^2$NVIDIA \\
  \texttt{\{pmolodyk3, jchoi843, yongchen\}@gatech.edu}
}

\begin{document}
\maketitle
\begin{abstract}
In recent years, point cloud generation has gained significant attention in 3D generative modeling. Among existing approaches, point-based methods directly generate point clouds without relying on other representations such as latent features, meshes, or voxels. These methods offer low training cost and algorithmic simplicity, but often underperform compared to representation-based approaches. In this paper, we propose \textbf{MFM-Point}, a multi-scale Flow Matching framework for point cloud generation that substantially improves the scalability and performance of point-based methods while preserving their simplicity and efficiency. Our multi-scale generation algorithm adopts a coarse-to-fine generation paradigm, enhancing generation quality and scalability without incurring additional training or inference overhead. A key challenge in developing such a multi-scale framework lies in preserving the geometric structure of unordered point clouds while ensuring smooth and consistent distributional transitions across resolutions. To address this, we introduce a structured downsampling and upsampling strategy that preserves geometry and maintains alignment between coarse and fine resolutions. Our experimental results demonstrate that MFM-Point achieves best-in-class performance among point-based methods and challenges the best representation-based methods. In particular, MFM-point demonstrates strong results in multi-category and high-resolution generation tasks.
\end{abstract}    
\begin{figure*}[t]
    \center
    \includegraphics[width=0.95 \linewidth]{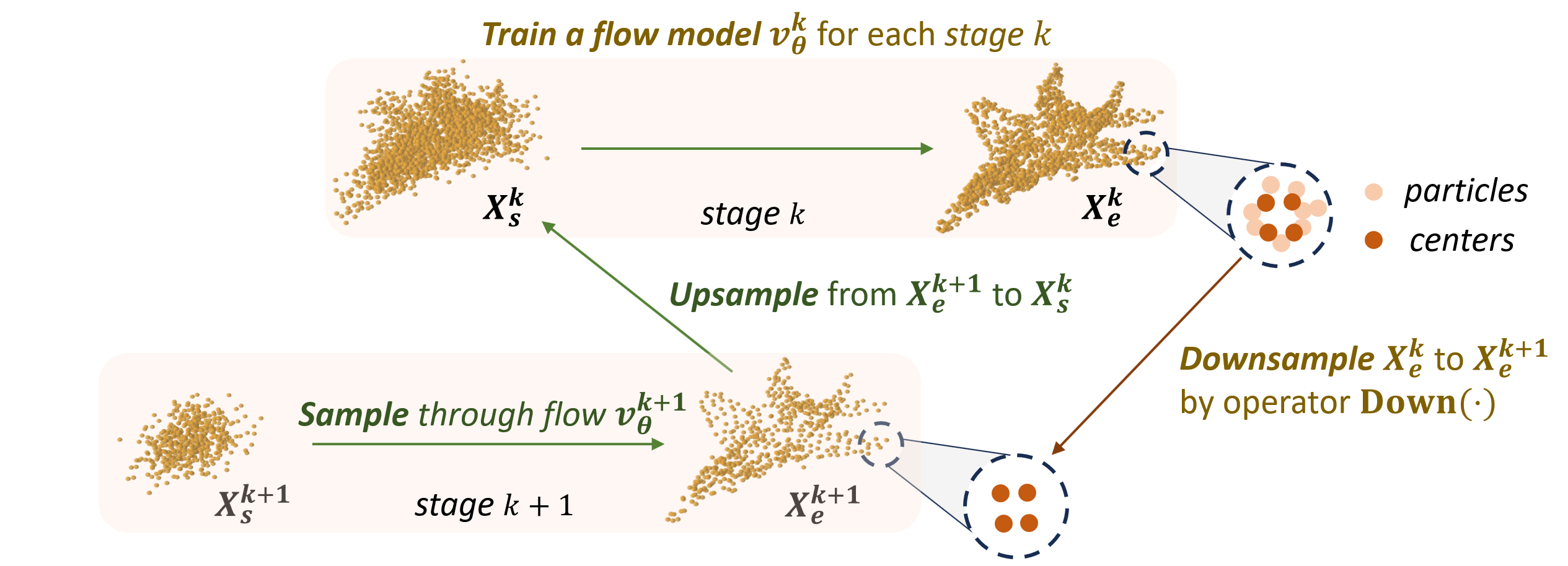}
    \caption{\textbf{Overview of the proposed multi-scale point cloud generation framework.}
The framework consists of two processes: (1) \textit{\textbf{\textcolor{darkbrown}{Training}}} (shown in brown) and (2) \textit{\textbf{\textcolor{darkgreen}{Inference}}} (shown in green). During training, we perform downsampling to obtain the coarse representation $X^{k+1}_e = \text{Down}(X^k_e)$, and train a flow model $v^k_\theta$ to transport samples from the distribution of $X^k_e$ at each stage $k$. During inference, we first sample points through the learned flow $v^{k+1}_\theta$ from $X^{k+1}_s$ to $X^{k+1}_e$, and then upsample the result to reconstruct $X^k_s$, the input for the next finer stage. We employ an \textit{equal-size K-means clustering} strategy as the downsampling operator $\text{Down}(\cdot)$, followed by a carefully designed upsampling procedure that ensures \textit{distributional alignment} between the upsampled samples and the corresponding fine-scale distribution.}
\label{fig:concept}
\end{figure*}

\section{Introduction}
3D point cloud generation is a fundamental problem in 3D generative modeling with broad applications in shape generation, 3D reconstruction, and autonomous robotics. Among various modeling paradigms, point-based methods \cite{psf, pvd, not-so-ot, setvae} have been extensively studied for their ability to directly generate 3D points without relying on intermediate representations such as latents, meshes, or voxels. 
These point-based approaches are appealing due to their low training cost, as they avoid expensive 3D voxelization and complex encoder–decoder architectures. They also offer algorithmic simplicity, making them both easy to implement and computationally efficient. This simplicity further facilitates the study of geometric priors and generative behavior, enabling the analysis of modeling principles without the confounding influence of heavy architectural components \citep{simpledm, simplerdm}.
Over the years, a variety of point-based generative frameworks have been explored, ranging from variational autoencoders (VAEs) \cite{kingma2013auto, setvae} and generative adversarial networks (GANs) \cite{gan, wu2016learning} to, more recently, normalizing flows \cite{pointflow, sanghi2022clip, klokov2020discrete}, autoregressive models \cite{sun2020pointgrow, ibing2023octree, kol2022rpg}, diffusion and Flow Matching (FM) approaches \cite{pvd, lion, psf, wfm, not-so-ot}. However, point-based methods often underperform compared to representation-based approaches \cite{lion, frepolad, dit-3d}, which can better capture global structures at the cost of greater model complexity and computational overhead.

To overcome these limitations, we propose MFM-Point, a novel multi-scale Flow Matching (FM) framework for 3D point cloud generation. While prior point-based methods primarily focused on improving pair-matching strategies and loss formulations \cite{wfm, not-so-ot}, our work takes an orthogonal direction by enhancing scalability and performance through a multi-scale generation framework. Specifically, MFM-Point generates point clouds in several stages  of increasing resolution in a coarse-to-fine manner, where an independent flow model is trained at each stage to generate samples at the corresponding resolution. 
This progressive generation strategy
enables MFM-Point to efficiently handle high-resolution and multi-category tasks. 
Each flow model is trained using the flow matching objective, which reduces generation to a simple regression problem. Consequently, our training pipeline remains conceptually simple and computationally efficient, while maintaining strong expressivity to tackle complex generation tasks.

A key challenge in our multi-scale design lies in constructing geometrically consistent coarse-to-fine generation trajectories, as illustrated in \cref{fig:concept}. This means that fine-grained stages should use the coarse-grained information learned from previous resolutions and further refine the point cloud by gradually increasing the fidelity while preserving the overall geometry learned in the early stages.
We discover that this can be achieved by using a structured point cloud upsampling and downsampling algorithm that can transfer the information from the low resolution point cloud to the next model when switching between stages during inference. To acheve this, we derive the theoretical requirements placed on the upsampling and downsampling algorithms used to align the point clouds between consecutive stages, and develop an algorithm that satisfies these requirements.


We establish the effectiveness of MFM-Point through extensive experiments on high-resolution and multi-category point cloud generation. We demonstrate that our framework is both highly scalable and capable of producing diverse high-fidelity samples. Furthermore, our comprehensive ablation studies show that these performance gains stem directly from the proposed multi-scale architecture, highlighting the benefits of our coarse-to-fine design in improving both quality and efficiency. Additionally, we demonstrate class-conditional point cloud generation to underscore the flexibility and generalizability of MFM-Point across different generative settings. Our main contributions are summarized as follows:
\begin{itemize}[topsep=2pt, leftmargin=*, itemsep=1pt]
    \item We propose MFM-Point, a scalable multi-scale Flow Matching framework for point cloud generation. Its coarse-to-fine generation paradigm enables generation in complex settings while preserving the algorithmic simplicity and efficiency of point-based models.
    \item We derive the theoretical requirements for cross-stage alignment, providing a principled mechanism for consistent multi-scale generation. We offer a specific implementation by designing new downsampling and upsampling operators that preserve local geometry, thereby improving overall generation performance. 
    \item Our method achieves best-in-class performance among point-based methods and comparable results to that of the best representation-based approaches, with particularly strong results in multi-category and high-resolution scenarios. Extensive ablation and further experiments validate the effectiveness and generalizability of our framework.
\end{itemize}

\section{Background}
\label{sec:background}
\paragraph{Flow Matching}
Flow matching models \citep{flowmatching, rectifiedFlow, instaflow} aim to learn a velocity field $v:[0,T]\times \mathbb{R}^d \rightarrow \mathbb{R}^d$ that transports a given source distribution $\mu$ to a target distribution $\nu$ through an ordinary differential equation (ODE).
Given any joint distribution $\gamma \in \Pi (\mu, \nu)$, where $\Pi$ is the joint distribution of distributions $\mu$ and $\nu$, the flow matching model \citep{flowmatching} minimizes a simple regression objective:
\begin{equation} \label{eq:fm}
    \mathbb{E}_{(x_0,x_1)\sim \gamma} \left[ \Vert v_\theta (t, x_t) - (x_1 - x_0) \Vert^2 \right],
\end{equation}
where $x_t = (1-t) x_0 + t x_1$ for $t\in [0,1]$.
Under mild regularity conditions, the optimal velocity field $v_\theta (t,\cdot)$ transports the source distribution $\mu$ to the target $\nu$ via the following ODE dynamics:
\begin{equation}
    \vspace{-5pt}
    \dot{X}_t = v_\theta (t,X_t), \quad X_0 \sim \mu.
\end{equation}
In generative modeling, we typically set the source distribution as a standard Gaussian, $\mu = \mathcal{N}(0, I)$, and the target distribution as the data distribution, $\nu = p_{\text{data}}$.
Then, the probability of $x_t$ given $x_1 \sim p_{\text{data}}$ is given by:
\begin{equation}
    x_t = t x_1 + (1-t) n \sim \mathcal{N} (tx_1, (1-t)^2 I),
\end{equation}
where $n\sim \mathcal{N}(0, I)$.
Similar to diffusion models \citep{ddpm, scoresde}, flow matching models rely on a simple regression objective, making them easy and stable to train.

\paragraph{Multi-Scale Flow Matching for Image Generation.}
Recent progress in flow matching has enabled multi-scale generation for image and video synthesis \citep{pyramidFM, edify}, where data are generated progressively from coarse to fine resolutions. 
Typically, multi-scale flow matching models partition the overall time interval $[0,1]$ into $K$ subintervals $[s_k, e_k]$ for $k = 0,\ 1,\dots, K{-}1$, and assign a flow model $v^k_\theta$ on the sub interval to each stage $k$. 
At stage $k$, the model learns to transport samples from a distribution of initial states $x^k_s$ to a distribution of terminal states $x^k_e$ through ODE dynamics:
\begin{align}\label{eq:initial-terminal-state1}
    \begin{split}
    x^k_{e} &= e_k \, \text{Down}^k(x_1) + (1-e_k) n, \\ 
    x^k_{s} &= s_k \, \text{Up}(\text{Down}^{k+1}(x_1)) + (1-s_k) n,
    \end{split}
\end{align}
where $x_1 \sim p_{\text{data}}$ denotes a real sample, $n \sim p_{\text{prior}}$ is the noise prior, and $\text{Down}^k(\cdot)$ is a downsampling operator with factor of $2^k$, and $\text{Up}(\cdot)$ is a simple upsampling operator. 
Each flow model $v^k_\theta$ is trained using the flow matching objective:
\begin{align}\label{eq:FM-objective}
    \mathbb{E}_{t \sim \mathcal{U}[s_k, e_k],\ (x^k_s, x^k_e)} 
    \left[ \left\| v^k_\theta(t', x_t) - (x^k_e - x^k_s) \right\|^2 \right],
\end{align}
where $t' = (t - s_k)/(e_k - s_k)$ and $x_t = (1 - t')x^k_s + t'x^k_e$. 
This objective encourages $v^k_\theta$ to transport the distribution of $x^k_s$ to that of $x^k_e$. 
During inference, samples are generated sequentially: the model $v^{k+1}_\theta$ transforms $x^{k+1}_s$ to $x^{k+1}_e$, and the resulting $x^{k+1}_e$ is then upsampled to serve as input $x^k_s$ for the next, finer stage. Repeating this process yields the final high-resolution output.

\section{Multi-scale Flow Matching for Point Cloud Generation}
\label{sec:method}

\paragraph{Overview} We follow the notations introduced in \cref{sec:background}, except that we denote point cloud data using the uppercase letter $X$ instead of the lowercase $x$. To extend the flow matching framework to the multi-scale point cloud setting, we further need to define the downsampling and upsampling operators, $\text{Down}(\cdot)$ and $\text{Up}(\cdot)$, for point clouds. 
We then follow the formulation in \cref{eq:initial-terminal-state1}, where the initial and terminal states at each stage $k$ are defined as:
\begin{align}\label{eq:interpolate}
    \begin{split}
    &X^k_s = s_k \, \text{Up}(X^{k+1}_1) + (1 - s_k) n,  \\
    &X^k_e = e_k \, X^{k}_1 + (1 - e_k) n, \ \ n \sim \mathcal{N}\left(0, \frac{1}{D^k}I\right), \\[-5pt]
    \end{split} 
\end{align}
where $X^{k}_1$ and $X^{k+1}_1$ denote the point clouds obtained by repeatedly downsampling the original data $X_1$ $(= X^0_1)$ as
\begin{align*}
    X^k_1 \!=\! \text{Down}^k (X_1), \ \ 
    X^{k+1}_1 \!=\! \text{Down}(X^k_1) \!=\! \text{Down}^{k+1}\! (X_1).
\end{align*}
As illustrated in \cref{fig:concept}, the flow model $v^k_\theta$ learns a continuous velocity field that transports the distribution of the input point cloud $X^k_s$ to that of the target point cloud $X^k_e$. 
During inference, the learned flow $v^{k+1}_\theta$ transports $X^{k+1}_s$ to $X^{k+1}_e$ via ODE flow, and the resulting $X^{k+1}_e$ is then upsampled to form the initial distribution of the next finer stage, $X^{k}_s$. 

\paragraph{Challenges} To develop such a multi-scale framework for point clouds, two key challenges must be addressed: 
\begin{enumerate}[label=(\roman*), topsep=3pt, leftmargin=17pt, itemsep=2pt]
  \item Unlike grid-structured data such as images, point clouds lack a regular topology, making it \textbf{nontrivial to define downsampling and upsampling operators} that preserve local and global spatial properties.
  \item The design of operators must ensure that \textbf{the upsampled distribution of $X^{k+1}_e$ aligns with the distribution of $X^{k}_s$} in \cref{eq:interpolate}, thereby enabling accurate generation in the subsequent stage. 
\end{enumerate}
In the following sections, we establish new training and inference algorithms that address these challenges through geometric-aware downsampling strategies and principled upscaling schemes tailored for point cloud data, ensuring both structural integrity and statistical alignment across stages.

\subsection{Downsampling \& Upsampling Operator} \label{sec:downsample}

\paragraph{Downsampling Operator}
To address \textbf{challenge (i)}, it is essential to establish a spatial correspondence between $X^k_s$ and $X^k_e$. 
As indicated in \cref{eq:interpolate}, the geometric coherence between $X^k_s$ and $X^k_e$ is governed by the relationship between $\text{Up}(X^{k+1}_1)$ and $X^k_1$. 
To ensure such consistency, we employ a clustering algorithm in which each point in $X^{k+1}_1$ serves as the \emph{center of a cluster} in $X^k_1$.
Next, to address \textbf{challenge (ii)}, we further enforce \emph{equal-sized clustering} when constructing $X^{k+1}_1$. This constraint is not merely a design convenience but a theoretical necessity: by ensuring that each cluster contributes equally to the reconstruction of $X^k_s$, it guarantees the \emph{distributional alignment} formalized in \cref{thm:inference}. Together, these principles yield a downsampling operator that preserves geometric integrity within each stage and statistical consistency across stages, both of which are indispensable for accurate and stable multi-scale point cloud generation. 
Further implementation details are provided in \cref{sec:training_inference}.

\paragraph{Implementation Details}
We apply K-means clustering with a constrained cluster size of $D$, where $D$ represents the downsampling ratio. 
Since K-means clustering has low convergence rate, we initialize the cluster centers using the farthest point sampling (FPS) algorithm, which is commonly used for generating sub-samples in point cloud data \citep{pointnetplusplus}. This initialization gives a more uniform distribution of the initial cluster centers, leading to faster convergence. 
Finally, we concatenate the resulting clusters to construct the downsampled point cloud (see line 6-7). Formally, $m$-th point $x^{k+1}_m$ of $X^{k+1}_1$ is the center of $\{x^k_{Dm}, x^k_{Dm+1}, \dots, x^k_{D(m+1)-1} \} \subset X^k_1$. Detailed algorithm is presented in \cref{alg:downsample}.

\paragraph{Upsampling Operator} We next describe the upsampling operator $\text{Up}(\cdot)$ in \cref{eq:interpolate}.
We simply upsample $X^{k+1}_1$ by replicating each point $c_m {\in} X^{k+1}_1$ $D$ times. Formally, we define the upsampling operation as $\text{Up}(X^{k+1}_1) {=} \text{Concat}(\left\{ C_m \right\}^M_{m=1})$ where each set $C_m$ is a $D$-times replica of $c_m$, i.e. $C_m {:=} \{c_m\}^D_{i=1}$.
\begin{algorithm}[t]
\caption{Training Scheme for stage $k$}
\label{alg:training}
\small
\begin{algorithmic}[1]
\Require Stage index $k$; time interval $[s_k, e_k]$; downsample ratio $D$;
         joint distribution $p_k$ of $(X^{k+1}_1, X^k_1)$; flow model $v^k_\theta$
\For{each iteration}
    \State Sample a batch $(X^{k+1}_1, X^k_1) \sim p_k$.
    \State Sample time $t \sim \mathcal{U}[0,1]$.
    \State Sample noise $n \sim \mathcal{N}\!\left(0,\, \tfrac{1}{D^k} I_{d_k}\right)$.
           with $d_k = N / D^{k}$.
    \State Compute boundary states:
          \vspace{-0.05in}
          \[
          X^k_{s} \!\gets\! s_k\mathrm{Up}(X^{k+1}_1) + (1\!-\!s_k)n, \ \
          X^k_{e} \!\gets\! e_k X^k_1 + (1\!-\!e_k)n.
          \]
          
    \State Compute intermediate state $X^k_t \!\gets\! (1\!-\!t)\,X^k_{s} + t\,X^k_{e}.$
          
    \State Train by minimizing
        \vspace{-0.05in}
          \[
          \mathcal{L}(\theta) =
          \big\| v^k_\theta(t, X^k_t)
          - \big(X^k_{e} - X^k_{s}\big) \big\|^2.
          \]
        \vspace{-0.2in}
\EndFor
\end{algorithmic}
\end{algorithm}

\subsection{Training \& Inference Schemes}\label{sec:training_inference}

\paragraph{Training Scheme}
For each stage $k$, we train a flow model $v^k_\theta$ which connects distribution of initial states $X^k_s$ to that of terminal states $X^k_e$. Following \cref{eq:FM-objective}, we train the flow model $v^k_\theta$ for each stage $k$ as follows:
\begin{align*} \label{eq:FM-objective-pc}
    \\[-0.3in]
    \mathbb{E}_{t \sim \mathcal{U}[s_k, e_k],\ (X^k_s, X^k_e)}\left[ \left\| v^k_\theta(t', X_t) - (X^k_e - X^k_s) \right\|^2 \right],\\[-0.3in]
\end{align*}
where $\quad t' = \frac{t - s_k}{e_k - s_k}$ and $X_t = (1 - t') X^k_s + t' X^k_e$. The precise algorithm is written in \cref{alg:training}.

\begin{figure*}[t]
    \centering
    \begin{minipage}{0.45\linewidth}
        \centering
        \includegraphics[width=0.9\textwidth]{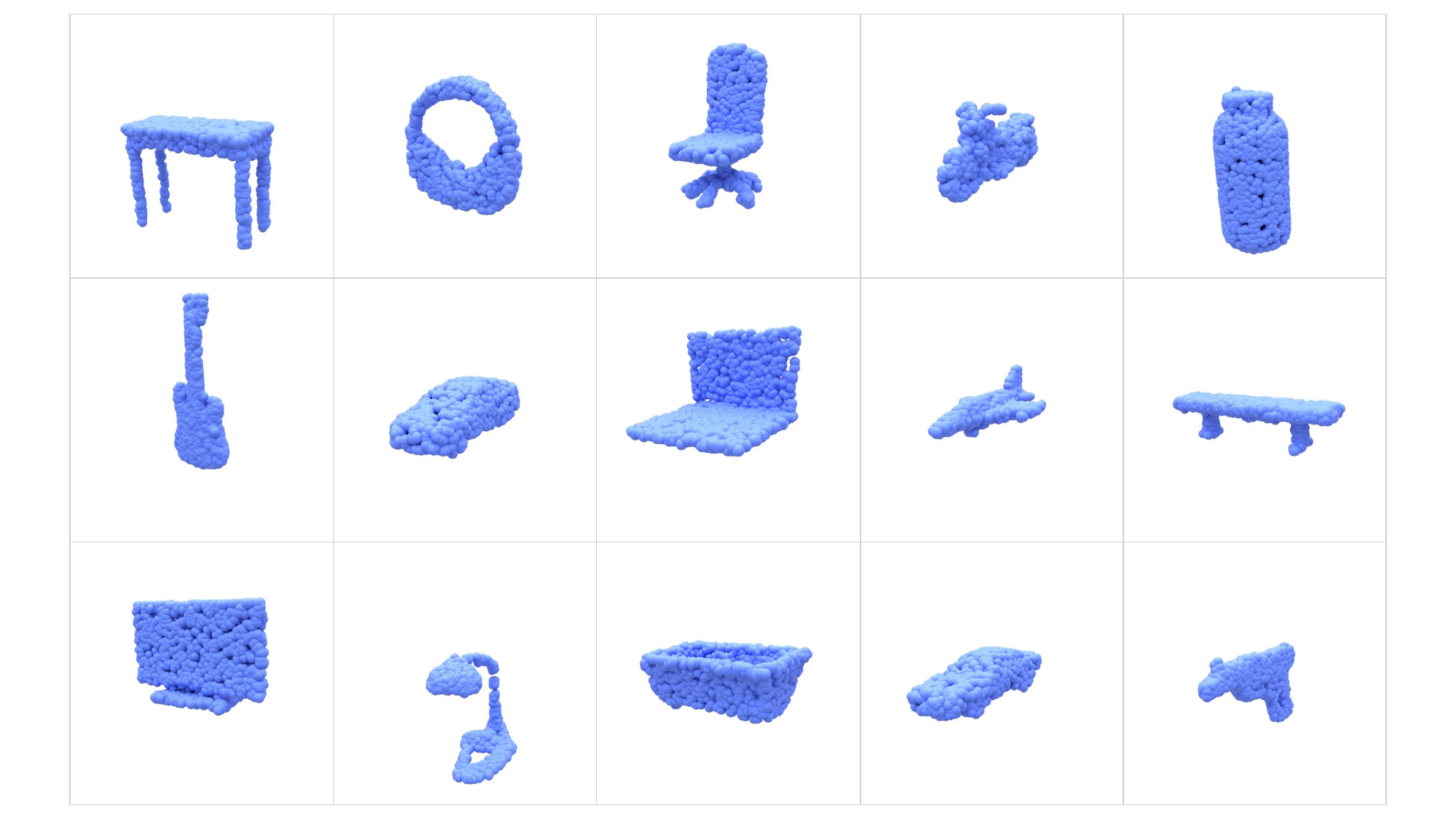}
        \vspace{-0.1in}
        \caption{Generated samples from our 2-stage model on the multi-category setting.}
        \label{fig:main-all-2048}
    \end{minipage}
    \hfill
    \begin{minipage}{0.45\linewidth}
        \centering
        \includegraphics[width=0.9\textwidth]{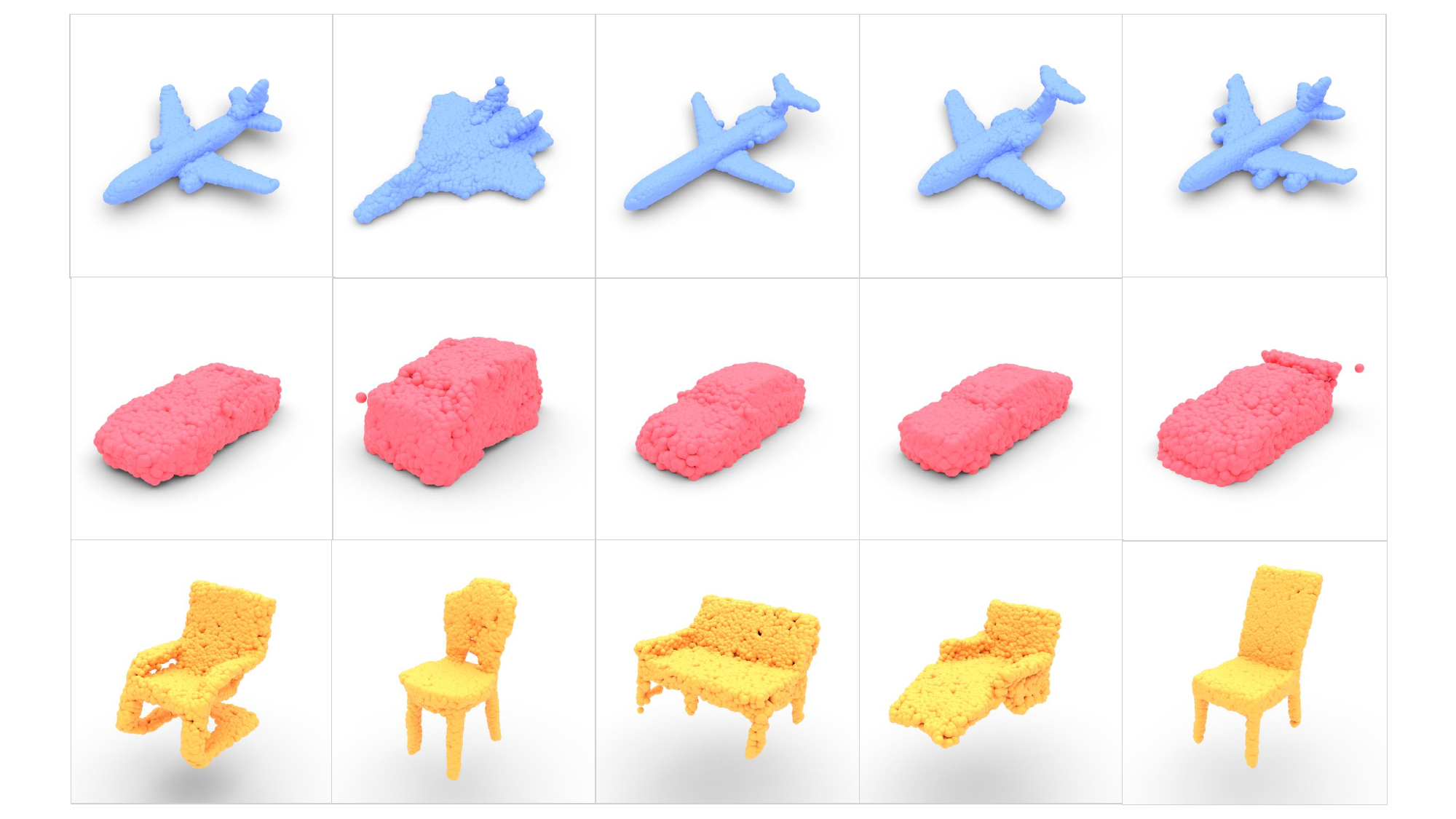}
        \vspace{-0.1in}
        \caption{Generated samples from our 2-stage model on the single-category setting.}
        \label{fig:main-single-8192}
    \end{minipage}
\end{figure*}

\paragraph{Distributional Alignment across stages}
The terminal state $X^{k+1}_e$ at stage $k{+}1$ and the initial state $X^k_s$ at stage $k$ are constructed according to \cref{eq:interpolate}, using our newly designed downsampling and upsampling operators. 
During inference,  the key objective is to accurately reconstruct the finer resolution distribution of $X^k_s$ given the coarser representation $X^{k+1}_e$, ensuring faithful distributional alignment across stages. Thanks to our newly developed downsample and upsample operators with \textit{equal-sized clustering}, the following theorem holds:
\begin{theorem} \label{thm:inference}
    Suppose $X^k_s$ and $X^{k+1}_e$ are defined as in \cref{eq:interpolate}, and $0 \leq e_{k+1} \leq s_k \leq 1$ for all $k\in \{0,1,\dots, K-1\}$. Then, we have
    \begin{equation}
    \setlength{\abovedisplayskip}{4pt}
    \setlength{\belowdisplayskip}{4pt}
        X^k_s - \tfrac{s_k}{e_{k+1}}\,\mathrm{Up}\!\left(X^{k+1}_e\right)
        \sim \mathcal{N}\!\left(0,\,\Sigma'\right),
    \end{equation}
    where the covariance matrix $\Sigma':= \mathrm{Diag}\!\big(\{\Sigma'_{D\times D}\}_{m=1}^M\big)$ is positive semi-definite block-diagonal with each block matrix $\Sigma'_{D\times D}$ given by
    \begin{align*}
        \\[-0.3in]
        \Sigma'_{D\times D}
        = \frac{(1-s_k)^2}{D^{k}} I_{D\times D}
          - \frac{s_k^2(1-e_{k+1})^2}{e_{k+1}^2 D^{k+1}}
            \mathbbm{1}_D \mathbbm{1}_D^\top.\\[-0.3in]
    \end{align*}
    Here, $\mathbbm{1}_D$ denotes the $D$-dimensional all-ones vector.
    In other words, the distribution of $X^{k}_s$ is equivalent to the distribution of
    \begin{equation} \label{eq:upsample_inference}
    \setlength{\abovedisplayskip}{5pt}
        \frac{s_k}{e_{k+1}} \text{Up}(X^{k+1}_e) + n' \ \ \ \text{ where } \ \ n' \sim \mathcal{N}(0, \Sigma').
    \end{equation}
\end{theorem}
\begin{remark}
    \cref{thm:inference} establishes that \textbf{the distribution of $X^k_s$ can be faithfully recovered from $X^{k+1}_e$ via \cref{eq:upsample_inference}}. An important property of this formulation is that the resulting covariance matrix $\Sigma'$ is \textit{positive semi-definite}, ensuring that it defines a valid Gaussian distribution and thus enabling noise sampling as $n' \sim \mathcal{N}(0, \Sigma')$. Notably, the noise variance $\frac{1}{D^k} I$ in \cref{eq:interpolate} is deliberately chosen to guarantee the positive semi-definiteness of $\Sigma'$. A detailed proof is provided in \cref{appen:proofs}. Furthermore, the following corollary presents a special case of \cref{eq:upsample_inference}, which admits a simple computation:
\end{remark}
\begin{corollary}
    Suppose $0\leq s_k < e_k \leq 1$. Then, the following holds:
    \begin{enumerate}[label=(\roman*), leftmargin=17pt, itemsep=2pt, topsep=5pt]
        \item Suppose $s_{k} = e_{k+1}$. Then, \cref{eq:upsample_inference} can be reduced as: 
        \begin{equation}
        \setlength{\abovedisplayskip}{4pt}
        \setlength{\belowdisplayskip}{4pt}
            \text{Up}(X^{k+1}_e) + \frac{1-s_k}{D^{k/2}} n \ \text{ where } \ n \sim \mathcal{N}(0, \bar{\Sigma}),
        \end{equation}
        where $\bar\Sigma = \text{Diag} \left(\{ I - \frac{1}{D} \mathbbm{1}_{D\times D} \}^M_{m=1} \right) $.
        \item Suppose $e_{k}=1$. Then, \cref{eq:upsample_inference} can be reduced as:
        \begin{equation}
        \setlength{\abovedisplayskip}{2pt}
\setlength{\belowdisplayskip}{2pt}
            s_k \text{Up}(X^{k+1}_e) + \frac{1-s_k}{D^{k/2}} n, \ \text{ where } \ n \sim \mathcal{N}(0, I).
        \end{equation} 
    \end{enumerate}
\end{corollary}

\begin{algorithm}[t]
\caption{Inference Scheme}
\label{alg:inference}
\small
\begin{algorithmic}[1]
\Require Time intervals $\{[s_k, e_k]\}^K_{k=1}$; flow models $\{v^k_\theta\}^K_{k=1}$; downsample ratio $D$; 
\State Sample $X^K_0 (= X^K_s) \sim \mathcal{N}(0, I)$.
\State  Compute $X^K_e$ from $X^K_s$ via flow model $v^K_\theta$.
\For{$k\in \{K-1, \dots, 0\}$}
    \State Sample noise $n' \sim \mathcal{N}\!\left(0,\, \Sigma' \right)$ as defined in \cref{eq:upsample_inference}.
    \State $X^k_s \leftarrow \frac{s_k}{e_{k+1}} \text{Up}(X^{k+1}_e) + n'.$
    \State Compute $X^{k}_e$ from $X^{k}_s$ via flow model $v^{k}_\theta$.
\EndFor \\
\Return $X^0_e$
\end{algorithmic}
\end{algorithm}

\paragraph{Inference Scheme}
Combining \cref{eq:upsample_inference} with standard flow generation, our generative procedure can be summarized as follows:
\begin{align*}
 X^K_s \!\to\! 
 \cdots 
 \!\to\!
X^{k+1}_s \xrightarrow[\text{$v^{k+1}_\theta$}]{}
X^{k+1}_e \xrightarrow[\text{\cref{eq:upsample_inference}}]{}
X^k_s \!\to\! \cdots \!\to\!
X^0_e, \\[-0.3in]
\end{align*}
starting from stage $K$ to 0, where $X^K_s (=X^K_0)  \sim \mathcal{N}(0, \frac{1}{D^K} I)$.
The precise algorithm is written in \cref{alg:inference}.

\paragraph{Preprocessing}
As shown in \cref{eq:interpolate}, the pair $(X^k_s, X^k_e)$ can be obtained through $(X^{k+1}_1, X^k_1)$ pair. By preprocessing a $X^k_1 = \text{Down}^k(X^0_1)$ for all $k=0,1,\dots, K-1$, we can directly draw a flow matching pair from the preprocessed dataset. Thus, instead of computing the clustering algorithm for every training iterations, we pre-compute it once before training, reducing the time burden of training.

\begin{figure*}[t]
    \centering
    \begin{minipage}{.67\linewidth}
    \centering
    \captionof{table}{
        Unconditional generation results on high-resolution point clouds (8192 and 15K points), evaluated using one-nearest neighbor accuracy (1-NNA) under Chamfer Distance (CD~$\downarrow$) and Earth Mover’s Distance (EMD~$\downarrow$). Results are reported for ShapeNet (\textit{Airplane}, \textit{Car}, \textit{Chair}) and Objaverse-XL (\textit{Furniture}). Bold numbers indicate the best overall performance, while underlined numbers denote results within a 1.0 margin of the best.
        } \label{tab:8K_results}
        \scalebox{0.85}{
        \begin{tabular}{l|cc|cc|cc|cc|cc}
        \toprule
        & \multicolumn{8}{c|}{\textbf{8192 points}} & \multicolumn{2}{c}{\textbf{15K points}} \\
        \midrule
        \textbf{Model}  & \multicolumn{2}{c|}{\textbf{Airplane}} & \multicolumn{2}{c|}{\textbf{Chair}} & \multicolumn{2}{c|}{\textbf{Car}} & \multicolumn{2}{c|}{\textbf{Furniture}} & \multicolumn{2}{c}{\textbf{Airplane}} \\
        & CD & EMD & CD & EMD & CD & EMD & CD & EMD & CD & EMD \\
        \midrule
        PVD  & 86.41 & 82.59 & 65.25 & 67.97 & 71.87 & 70.88 & 59.37 & 62.79 & 83.08 & 74.69 \\
        FM   & 82.09 & 70.12 & \cellhi\textbf{60.27} & 	\underline{58.83} & 66.05 & 62.50 & 73.63 & 74.80 & 86.17 & 68.39 \\
        Ours & \cellhi\textbf{69.62} & \cellhi\textbf{61.48} & \underline{60.57} & \cellhi\textbf{57.60} & \cellhi\textbf{60.22} & \cellhi\textbf{56.39} & \cellhi\textbf{57.90} & \cellhi\textbf{56.50} & \cellhi\textbf{74.81} & \cellhi\textbf{65.06}  \\
        \bottomrule
        \end{tabular}}
    \end{minipage}
    \vspace{5mm}
    \hfill
    \begin{minipage}{.31\linewidth}
    \centering
    \captionof{table}{
        Unconditional multi-category generation results on ShapeNet, evaluated over 3-categories (\textit{Airplane}, \textit{Car}, \textit{Chair}) and the full set of 55 categories. Bold numbers indicate the best overall performance.
        } \label{tab:multiclass_results}
        \scalebox{0.85}{
        \begin{tabular}{l|cc|cc}
        \toprule
        \textbf{Model}  & \multicolumn{2}{c|}{\textbf{3-Categories}} & \multicolumn{2}{c}{\textbf{All}} \\
        & CD & EMD & CD & EMD  \\
        \midrule
        PVD & 62.57 & 56.76 & 58.93 & 56.54 \\
        LION & - & - & 58.25 & 57.75 \\
        Ours & \cellhi\textbf{60.81} & \cellhi\textbf{55.88} & \cellhi\textbf{57.12} & \cellhi\textbf{54.12} \\
        \bottomrule
        \end{tabular}}
    \end{minipage}
    
    \begin{minipage}{1\linewidth}
        \centering
        \captionof{table}{
        Unconditional generation results on the \textit{Airplane}, \textit{Car}, and \textit{Chair} categories at a resolution of 2048 points. We report one-nearest neighbor accuracy (1-NNA) under Chamfer Distance (CD~$\downarrow$) and Earth Mover’s Distance (EMD~$\downarrow$). Bold numbers indicate the best overall performance within each \textit{class}.
        } \label{tab:main_results}
        \scalebox{0.84}{
        \begin{tabular}{c|c|c|cc|cc|cc}
        \toprule
        \textbf{Class} & \textbf{Model} & \textbf{Method} & \multicolumn{2}{c|}{\textbf{Airplane}} & \multicolumn{2}{c|}{\textbf{Chair}} & \multicolumn{2}{c}{\textbf{Car}}  \\
        & & & CD ($\downarrow$) & EMD ($\downarrow$)& CD ($\downarrow$) & EMD ($\downarrow$)& CD ($\downarrow$) & EMD ($\downarrow$)  \\
        \midrule
        \multirow{13}{*}{Point-based} & \multirow{4}{*}{Flow Matching} & \textbf{MFM-point} (Ours) & \textbf{65.36} & \cellhi\textbf{57.21} & \textbf{54.92} & \textbf{53.25} & 57.23 & \cellhi\textbf{47.87} \\
        & & PSF \citep{psf} & 71.11 & 61.09 & 58.92 & 54.45 & 57.19 & 56.07 \\
        & & WFM \citep{wfm} & 73.45 & 71.72 & 58.98 & 57.77 & 56.53 & 57.95 \\
        & & NSOT \citep{not-so-ot} & 68.64 & 61.85 & 55.51 & 57.63 & 59.66 & 53.55 \\
        \cmidrule(lr){2-9}
        & \multirow{3}{*}{Diffusion} & DPM~\citep{dpm} & 76.42 & 86.91 & 60.05 & 74.77 & 68.89 & 79.97 \\
        & & PVD~\citep{pvd} (N=1000) & 73.82 & 64.81 & 56.26 & 53.32 & \textbf{54.55} & 53.83 \\
        & & PVD-DDIM~\citep{pvd} (N=100) & 76.21 & 69.84 & 61.54 & 57.73 & 60.95 & 59.35 \\
        \cmidrule(lr){2-9}
        & \multirow{6}{*}{Others} & 1-GAN~\citep{1-GAN} & 87.30 & 93.95 & 68.58 & 83.84 & 66.49 & 88.78 \\
        & & PointFlow~\citep{pointflow} & 75.68 & 70.74 & 62.84 & 60.57 & 58.10 & 56.25 \\
        & & DPF-Net~\citep{dpf-net} & 75.18 & 65.55 & 62.00 & 58.53 & 62.35 & 54.48 \\
        & & SoftFlow~\citep{softflow} & 76.05 & 65.80 & 59.21 & 60.05 & 64.77 & 60.09 \\
        & & SetVAE~\citep{setvae} & 75.31 & 77.65 & 58.76 & 61.48 & 59.66 & 61.48 \\
        & & ShapeGF~\citep{shapegf} & 80.00 & 76.17 & 68.96 & 65.48 & 63.20 & 56.53 \\
        \midrule
        \multirow{2}{*}{Latent-based} & \multirow{2}{*}{LDM} & LION~\citep{lion} & 67.41 & \textbf{61.23} & 53.70 & \textbf{52.34} & 53.41 & 51.14 \\
        & & FrePoLad~\citep{frepolad} & \textbf{65.25} & 62.10 & \textbf{52.35} & 53.23 & \textbf{51.89} & \textbf{50.26} \\
        \midrule
        \multirow{3}{*}{Others} & Voxel & NWD~\citep{nwd} & 59.78 & 53.84 & 56.35 & 57.98 & 61.75 & 58.54 \\
        & Neural fields & 3DShape2VecSet~\citep{3dshape2vecset} &  62.75 & 61.01 & 54.06 & 56.79 & 86.85 & 80.91 \\
        & Latent-Voxel & DiT-3D \citep{dit-3d} & \cellhi \textbf{62.35} & \textbf{58.67} & \cellhi\textbf{49.11} & \cellhi\textbf{50.73} & \cellhi\textbf{48.24} & \textbf{49.35}\\
        \bottomrule
        \end{tabular}}
    \end{minipage}
\end{figure*}

\section{Related Works}
\paragraph{Flow Matching for Point Cloud Generation} 
Flow Matching \citep{flowmatching, stochasticinterp} (FM) has been widely adopted in image generation tasks, and recent studies have extended its application to the point cloud domain \citep{psf, wfm, not-so-ot}. A fundamental challenge in applying generative models to point clouds stems from their permutation invariance, which makes it difficult to define meaningful correspondences and learn coherent transformation trajectories. To address this, several recent works have proposed methods for improving point correspondences to enhance FM-based training. Specifically, WFM \citep{wfm} and NSOT \citep{not-so-ot} attempt to reorder the prior noise samples to better align with the data distribution, thereby yielding improved ODE trajectories. However, these approaches incur significant computational overhead during training and result in limited performance gains. PSF \citep{psf} proposes an iterative reflowing procedure to refine FM trajectories by progressively straightening them, following the rectified flow scheme \citep{rectifiedFlow}. Although this method improves trajectory quality, it also requires multiple rounds of training, making it computationally expensive with only modest improvements in generation quality. Several methods \citep{psf, pvd, qi2016volumetric} use a technique called voxelization to convert point clouds to a permutation-invariant volumetric representation.

\paragraph{Point Cloud Generation}
A diverse range of generative models has been applied to point cloud generation, including point-based, latent-based, and voxel-based approaches, etc. Point-based methods directly generate point clouds without relying on intermediate representations such as voxels or meshes. The variational autoencoder (VAE)~\citep{kingma2013auto} was adapted for point clouds in \citep{setvae}, while several approaches~\citep{1-GAN, chen2019learningimplicitfieldsgenerative} leverage the generative adversarial network (GAN) framework~\citep{gan}. 
Prior to the advent of Flow Matching~\citep{flowmatching}, normalizing flows~\citep{norm-flow, kingma2018glow} were also explored for point cloud generation, most notably in \citep{pointflow, sanghi2022clip, klokov2020discrete}. 
More recently, denoising diffusion models~\citep{dpm, pvd} have achieved strong empirical performance in this setting. Latent-based methods \citep{lion, frepolad} adopt the latent diffusion model framework~\citep{ldm}, operating in a compact latent space to improve scalability and expressivity. However, these methods require complex training procedures, typically involving two or more encoder–decoder architectures and training distinct diffusion models for each latent representation. Voxel-based methods~\citep{nwd, trellis, lt3sd, dit-3d} have been developed for large-scale or scene-level generation, representing 3D shapes as volumetric grids. While voxel representations simplify convolutional processing, they often incur substantial memory and computational costs. Other representation paradigms, such as neural field–based models~\citep{sdfusion, 3dshape2vecset} have also been actively explored for 3D generation tasks.

\paragraph{Multi-scale Generation}
Several methods \cite{progan, vahdat2020nvae, gu2023matryoshka, lai2017deep} have explored multi-scale training to improve the scalability to large-scale datasets for images and videos. These include approaches based on diffusion models \citep{ho2022cascaded, gu2023matryoshka}, flow matching \citep{pyramidFM}, and autoregressive generation \citep{tian2024visual-var}. Notably, multi-resolution diffusion and flow matching have been shown to be mathematically equivalent, as discussed in \citep{edify}. However, all of these methods rely on the inherent upsampling and downsampling operations available in image domains, and are therefore not directly applicable to point clouds.

\section{Experiments}
\label{sec:experiments}
In this section, we conduct a comprehensive evaluation and analysis of our method. 
In \cref{sec:main-results}, we evaluate MFM-point on standard unconditional point cloud generation benchmarks. MFM-point demonstrates substantial performance gains over existing point-based methods on more challenging tasks such as multi-category generation and high-resolution point cloud synthesis.
As  presented in \cref{sec:ablation}, we conduct ablation studies to investigate the core design choices, including our downsampling strategy, and the transition time boundaries between stages. Furthermore, we demonstrate the versatility of our approach on the class-conditional point cloud generation task. Moreover, ablation on the number of stages is provided in \cref{append:additional-results}. More qualitative results are provided in \cref{appen:qualitative}.

\subsection{Unconditional Point Cloud Generation} \label{sec:main-results}

\paragraph{Experimental Settings}
We evaluate our model on standard unconditional point cloud generation benchmarks using the ShapeNet \citep{shapenet} and Objaverse-XL \citep{objaverse-xl} datasets. 
For ShapeNet experiments, we conduct single category and multi-category experiments. 
For the single-category setting, we focus on three representative object classes, \textit{Airplane}, \textit{Car}, and \textit{Chair}, and evaluate performance at multiple resolutions: 2048 points as the standard resolution and 8192 and 15K points for high-resolution generation. For the multi-category setting, we perform unconditional generation on both a 3-class subset (\textit{Airplane}, \textit{Car}, \textit{Chair}) and the full set of 55 ShapeNet categories. For Objaverse-XL, we preprocess the \emph{furniture} subset into point clouds and evaluate our model at a resolution of 8192 points. Unless otherwise stated, we train MFM-point with a two-stage model ($K{=}2$) with the stage transition boundary set to $t^0_s{=}0.6$ and $t^1_e{=}1$. Additional implementation details are provided in \cref{appen:implementation-details}.

\paragraph{Metrics \& Baselines} Following prior work~\citep{psf, pvd}, we adopt one-nearest neighbor accuracy (1-NNA) using Chamfer Distance (CD) and Earth Mover’s Distance (EMD) as similarity measures. Our main baselines are point-based approaches that employ diffusion or flow models, including PVD~\citep{pvd}, PSF~\citep{psf}, NSOT~\citep{not-so-ot}, and WFM~\citep{wfm}. 
We also compare with strong benchmarks such as LION~\citep{lion}, FrePoLad~\citep{frepolad}, and DIT-3D \citep{dit-3d}. 

\paragraph{High Resolution Point Cloud Generation}

We evaluate our model on high-resolution point clouds containing 8192 and 15K points, a particularly challenging setting that demands preserving both diversity and fine-grained geometric details. As shown in \cref{tab:8K_results}, our method achieves substantial improvements over existing baselines. Our method consistently \textbf{outperforms PVD and FM across all benchmarks}, except for the \textit{Chair} category in EMD, where the difference remains within 0.4. These results demonstrate the strong generalization ability of MFM-Point in high-resolution generation, where maintaining both diversity and fidelity is particularly challenging. Qualitative results in \cref{fig:main-all-2048}, \cref{fig:main-single-8192}, and \cref{appen:qualitative} further confirm this trend, showing that MFM-Point generates diverse and refined point clouds. We attribute these gains to our multi-scale generation framework: \textbf{the coarse stage enables exploration of diverse global structures, while subsequent finer stages progressively refine geometry}, enhancing local detail while maintaining global structure.

\paragraph{Multi-Category Point Cloud Generation}
We further evaluate MFM-Point in the multi-category setting, which is particularly challenging due to the structural variability across object classes. 
As shown in \cref{tab:multiclass_results}, MFM-Point not only outperforms point-based baselines such as PVD~\citep{pvd} but also surpasses LION~\citep{lion}, a latent diffusion model that relies on complex encoder–decoder architectures to learn representations of point cloud. Our method achieves Chamfer Distance (CD) and Earth Mover’s Distance (EMD) scores of 57.12 and 54.12, respectively, on the full 55-class benchmark, exceeding all comparisons. Remarkably, MFM-Point, despite being a purely point-based framework, outperforms latent diffusion models such as LION~\citep{lion}, demonstrating that our simple multi-scale design can rival and even surpass more representation-heavy generative paradigms. These findings highlight the scalability and strong generalization ability of MFM-Point across diverse and heterogeneous point cloud distributions.

\paragraph{Single Category Point Cloud Generation}
We evaluate MFM-Point on standard single-category benchmarks, as summarized in \cref{tab:main_results}. Overall, MFM-Point achieves the best performance among point-based approaches across most of the metrics. 
Notably, DiT-3D~\citep{dit-3d}, the strongest method among our comparisons, achieves the best CD scores. However, MFM-Point attains superior EMD results for the \textit{Airplane} and \textit{Car} categories, achieving the best scores among all comparisons. This finding is particularly interesting because both DiT-3D and latent diffusion models adopt encoder–decoder architectures, with DiT-3D further leveraging voxelized representations and large transformer backbones. \textbf{Despite its much simpler point-based design, MFM-Point attains competitive performance}, highlighting the effectiveness of our multi-scale flow matching framework. In addition, \textbf{our model trains substantially faster than LION \citep{lion} and DiT-3D \citep{dit-3d}}, requiring only 14 hours compared to more than 40 hours for these methods.


\begin{figure*}
    \centering
    \begin{minipage}{0.49\linewidth}
        \centering
        \includegraphics[width=0.9\textwidth]{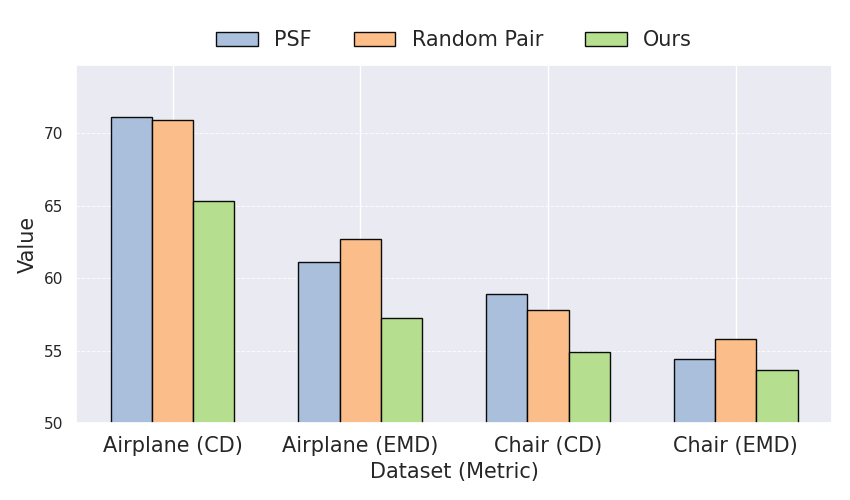}
        \vspace{-10pt}
        \caption{Comparison between PSF, \textit{Random Pair}, and Ours on Airplane and Chair dataset. \vspace{-2mm}
        }
        \label{fig:dnsample}
    \end{minipage}
    \hfill
    \begin{minipage}{0.49\linewidth}
        \centering
        \includegraphics[width=1\textwidth]{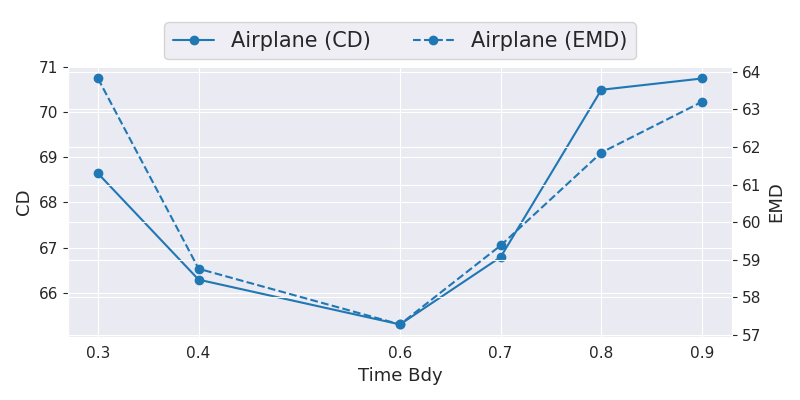}
        \vspace{-20pt}
        \caption{Ablation studies on time boundary $t^0_s$ for finer stage (Stage 0) on Airplane dataset.
        \vspace{-2mm}
        }
        \label{fig:time_bdy}
    \end{minipage}
    \hfill
    \begin{minipage}{0.485\linewidth}
        \centering
        \includegraphics[width=0.9\textwidth]{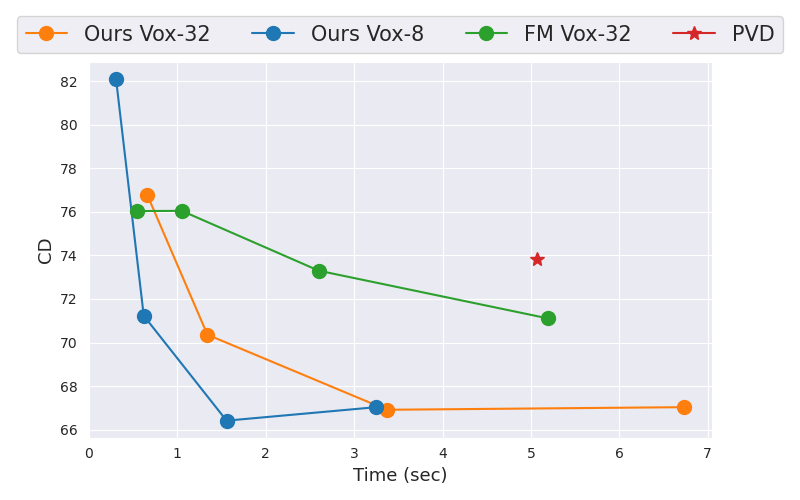}
        \vspace{-10pt}
        \caption{CD with respect to the inference time on FM and Ours in large (Vox-32) and small (Vox-8) model architecture. \vspace{-2mm}
        }
        \label{fig:time_to_cd}
    \end{minipage}
    \hfill
    \begin{minipage}{0.485\linewidth}
            \centering
            \includegraphics[width=0.85\textwidth, trim=3cm 0cm 3cm 0cm, clip]{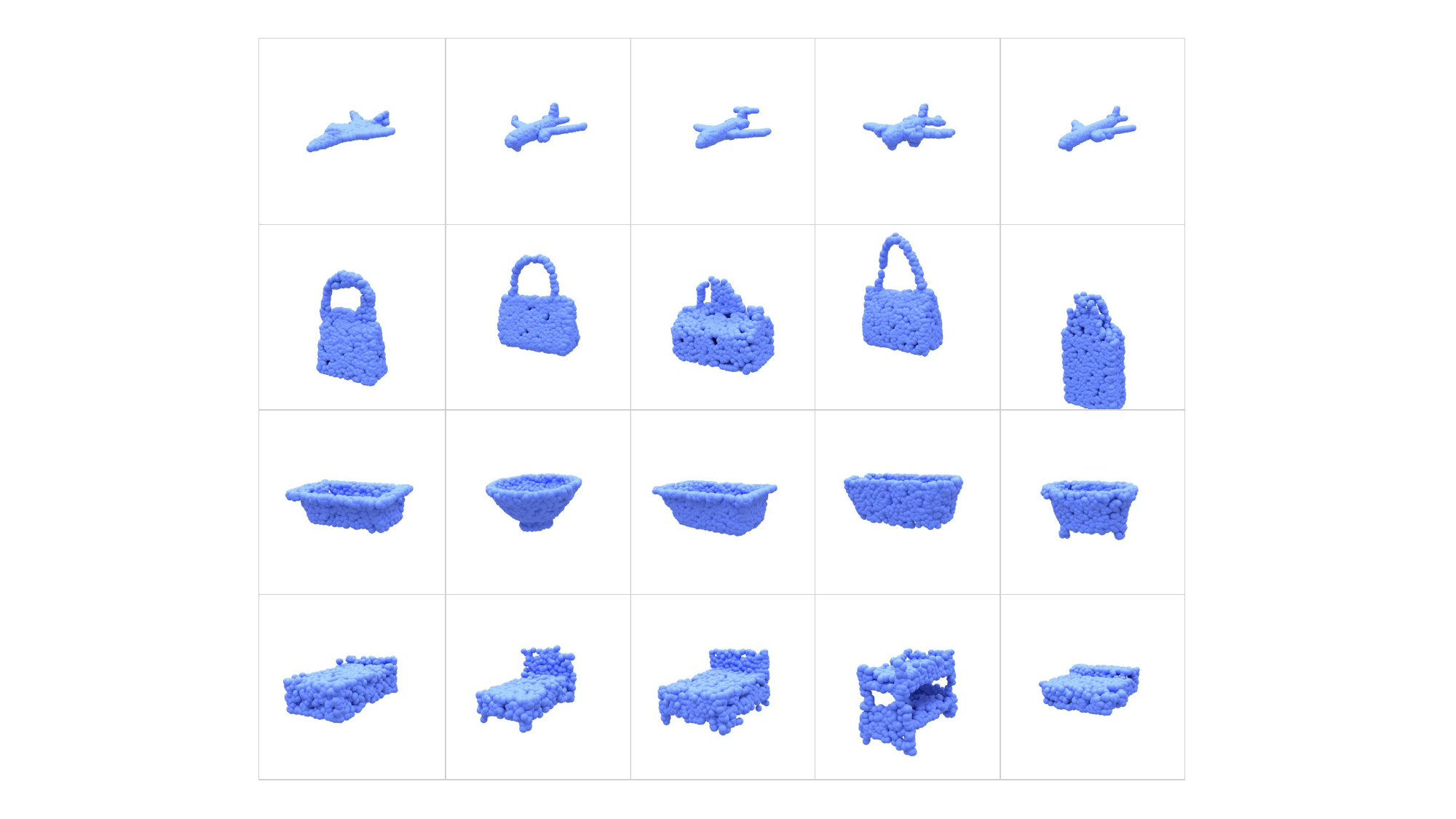}
            \vspace{-10pt}
            \caption{Class-conditional samples from our our two-stage model. Each row shows generations for Airplane, Bag, Bathtub, and Bed.
            }
            \label{fig:main-condition}
    \end{minipage}
    \vskip -0.1in
\end{figure*}

\subsection{Further Analysis} \label{sec:ablation}

\paragraph{Effect of Downsampling Operator}
We further investigate the impact of our geometry-aware downsampling design through an ablation study that isolates this component. 
To this end, we construct a baseline, denoted as \textit{Random Pair}, in which the downsampling operator $\text{Down}(\cdot)$ is defined as follows: 
given input data $X^0_1$, it is first randomly permuted and then partitioned into $D$ groups for average pooling, thereby destroying the spatial coherence that our geometric clustering enforces. 
This operation removes the correspondence between the upsampled coarse representation $X^0_s := \text{Up}(X^1_e)$ and the target $X^0_e$, breaking geometric consistency within each stage. 
As shown in \cref{fig:dnsample}, the \textit{Random Pair} baseline provides no improvement over the standard FM model, indicating that multi-scale modeling alone offers no benefit without geometric alignment. 
In contrast, MFM-Point achieves a clear performance gain, demonstrating that preserving geometric structure through our downsampling operator is crucial for producing coherent and high-fidelity point clouds. This highlights the need for a geometry-aware downsampling operator in multi-scale point cloud generation.


\paragraph{Effect of transition time boundaries} 
In our two-stage training setting, we investigate how the time point at which the high-resolution stage begins affects performance. Specifically, we train high-resolution models with thresholds ranging from $0.3$ to $0.9$ in increments of $0.1$ and evaluate their performance accordingly. The results for the airplane class are presented in \cref{fig:time_bdy}. Note that setting the threshold to $0$ effectively reduces the 2-stage model to standard FM in the high-resolution setting, as all low-resolution information is omitted. We observe that the generation quality reaches a sweet spot when the value lies in the range of approximately 0.5–0.7, while the performance degrades as it moves toward either extreme.

\paragraph{Inference Time}
We analyze the trade-off between inference time and output quality by varying the number of function evaluation (NFE) for inference. Specifically, we measure the wall-clock time required to generate a single point cloud by averaging over 256 samples, processed in batch size of 64. We compare three variants of flow matching algorithms: (1) single-stage flow matching applied at the highest resolution, (2) a two-stage implementation of our model (Vox-32), and (3) a two-stage model that employs a \textit{small} model (Vox-8). Note that the description of exact model architectures, including \textit{small} model, is described in Appendix \ref{appen:implementation-details}. Shown in \cref{fig:time_to_cd}, our model achieves superior performance compared to FM baseline while also requiring significantly less inference time. Notably, incorporating the smaller model in the low-resolution stage does not degrade performance and yields additional speed improvements.
\vspace{-2mm}
\paragraph{Conditional Generation}
Finally, we present qualitative results for class-conditional generation on the full dataset. Intuitively, \textit{Stage 0}, the finer resolution stage, is responsible for enhancing fidelity based on the coarse shape generated by \textit{Stage 1}. As such, it naturally preserves the class and overall structure provided by the coarser stage. Leveraging this insight, we train only the coarse-scale flow model (Stage 1) with class conditioning, while using the original unconditional model for the finer-scale flow model (Stage 0). As shown in \cref{fig:main-condition}, most generated samples successfully retain their class identity, even without explicit class guidance at Stage 0. This supports the effectiveness of our multi-scale framework in \textbf{decomposing generation into shape modeling and fidelity refinement}, and provides insight into why our method performs well in practice.

\section{Conclusion}
\label{sec:conclusion}
We presented MFM-point, a multi-scale flow matching framework for point cloud generation. We carefully designed downsampling and upsampling algorithms that meet two essential objectives: preserving geometric structure within each stage, and ensuring distributional alignment between coarse and fine representations to support consistent multi-scale inference. Through extensive experiments, we demonstrated that MFM-point consistently outperforms point-based methods, achieves performance comparable to recent competitive methods, and exhibits particularly strong results in high-resolution and multi-category settings. This highlights both the scalability and efficiency of our framework. Beyond benchmark performance, we conducted detailed analyses of the multi-scale architecture, including ablation studies on downsampling strategies. We further illustrated the flexibility of MFM-point through conditional generation experiments, underscoring its potential for broader applicability. A primary limitation of our work is that, although MFM-point substantially advances point-based methods, there are some datasets where the strongest representation-based methods show better performance, indicating that further enhancements may be necessary. Our work could also benifit from additional experiments on large-scale generation tasks such as scene-level synthesis. Extending our multi-scale framework to these more demanding settings represents an important direction for future research.
{
    \small
    \bibliographystyle{ieeenat_fullname}
    \bibliography{main}
}

\clearpage
\setcounter{page}{1}
\maketitlesupplementary

\section{Broader Impacts}\label{appen:broader}
Our approach improves the quality of point cloud generation and enhances scalability to high-resolution data. These advancements have the potential to contribute positively to society by supporting innovations in areas such as medical imaging, scene generation, and autonomous vehicle systems—all of which leverage point cloud data and stand to benefit from improved generative modeling techniques. At the same time, it is essential to acknowledge the potential risks associated with deploying synthetic data in real-world settings. Careful validation and responsible deployment practices are necessary to ensure safe and reliable outcomes when applying these models in physical environments.

\section{Proofs}\label{appen:proofs}
\paragraph{Proof of Theorem \ref{thm:inference}} 

\begin{proof}
    Let $X^0_1 \sim p_{\text{data}}$ be arbitrarily given data. By the definition, $X^k_s$ is defined as follows: 
    \begin{align*}
        &X^{k+1}_1 = \text{Down}^{k+1}(X^0_1), \ \ X^k_s = s_k \text{Up}(X^{k+1}_1) + (1-s_k),
    \end{align*}
    where  $n\sim \mathcal{N}\left(0, \frac{1}{D^k} I\right)$.
    Therefore, 
    \begin{equation}
        X^k_s \sim \mathcal{N}\left(s_k \text{Up}(X^{k+1}_1), \frac{(1-s_k)^2}{D^{k}} I \right).
    \end{equation}
    By definition of $X^{k+1}_e$, we can easily obtain $\text{Up}(X^{k+1}_e)$ as follows:
    \begin{equation}
        \text{Up}(X^{k+1}_e) = e_{k+1} \text{Up}(X^{k+1}_1) + (1-e_{k+1}) \text{Up}(n'), 
    \end{equation}
    where $n'\sim \mathcal{N}\left(0, \frac{1}{D^{k+1}}I\right)$. In other word, the distribution of $\text{Up}(X^{k+1}_e)$ can be written as follows:  
    \begin{equation}\label{eq:derive1}
        \text{Up}(X^{k+1}_e) \sim \mathcal{N}\left(e_{k+1} \text{Up}(X^{k+1}_1), \frac{(1-e_{k+1})^2}{D^{k+1}} \Sigma \right),
    \end{equation}
    where $\Sigma$ is the block diagonal matrix with blocks of size $D\times D$ filled with ones, i.e., $\Sigma = \text{Diag}\left( \{\mathbbm{1}_{D}\mathbbm{1}^T_{D} \}^M_{i=1} \right)$. 
    Thus, 
    \begin{equation}\label{eq:derive2}
        \frac{s_k}{e_{k+1}}\text{Up}(X^{k+1}_e) \sim \mathcal{N}\left(s_k \text{Up}(X^{k+1}_1), \frac{s^2_k (1-e_{k+1})^2}{ e^2_{k+1} D^{k+1}} \Sigma \right).
    \end{equation}
    Note that the distribution of $X^k_s$ and the distribution of $\frac{s_k}{e_{k+1}} \text{Up}(X^{k+1}_e)$ are both Gaussian that shares the same mean value. Therefore, the variance of $X^k_s - \frac{s_k}{e_{k+1}} \text{Up} (X^{k+1}_e)$ can be easily derived by taking subtraction between two variance matrices of Eq. \ref{eq:derive1} and Eq. \ref{eq:derive2}. Finally, we can obtain the following result:
\begin{equation}
    X^k_s - \frac{s_k}{e_{k+1}}\text{Up}(X^{k+1}_e) \sim \mathcal{N}(0, \Sigma'),
\end{equation}
where $\Sigma' = \text{Diag} \left(\{{\Sigma'}_{D\times D}\}^M_{m=1} \right),$
\begin{equation} 
     \quad {\Sigma'}_{D\times D} = \frac{(1-s_k)^2}{D^{k}} I_{D\times D} - \frac{s^2_k (1-e_{k+1})^2}{ e^2_{k+1} D^{k+1}} \mathbbm{1}_{D\times D}.
\end{equation}
Thus, the distribution of 
\begin{equation}
    \frac{s_k}{e_{k+1}} \text{Up}(X^{k+1}_e) + n' \text{ where } n' \sim \mathcal{N}(0, \Sigma'),
\end{equation}
is equivalent to the distribution of $X^{k}_s$.
Note that the sum of the row of $\Sigma'$ is non-negative whenever $e_{k+1} \geq s_k$, hence, $\Sigma'$ is positive definite matrix.
\end{proof}

\section{Implementation Details} \label{appen:implementation-details}

\paragraph{Downsampling Operator} The pseudo code for downsampling operator $\text{Down}(\cdot)$ is provided below:

\begin{algorithm}[h]
\caption{Downsampling Operator}
\small
\begin{algorithmic}[1]
\Require Point cloud $X^k_1$ with $N$ points; downsample ratio $D$; set $M \gets N/D$.
\State $\{c_m\}_{m=1}^M \gets \mathrm{FPS}(X^k_1)$
\Repeat
    \State $\{X_m\}_{m=1}^M \gets \mathrm{ConstrainedKMeans}(X^k_1, \{c_m\}_{m=1}^M)$
    \State $c_m \gets \frac{1}{D}\!\sum_{i=1}^{D} x_i^m,\ \forall\, m \in \{1,\dots,M\}$
\Until{converged}
\State $X^{k+1}_1 \gets \mathrm{Concat}\!\big(\{c_m\}_{m=1}^M\big)$
\State $X^{k}_1 \gets \mathrm{Concat}\!\big(\{X_m\}_{m=1}^M\big)$
\State \Return $(X^{k+1}_1, X^k_1)$
\end{algorithmic}
\label{alg:downsample}
\end{algorithm}

\paragraph{ShapeNet Preprocessing}
We run the downsampling algorithm on the ShapeNet dataset ahead of training. This only needs to be done once. We preprocessed the dataset with a downsampling ratio of 4. For every point cloud in the dataset, we run the downsampling algorithm and store the resulting low-resolution point cloud and the original in a structured way, ensuring that the points in the high-resolution version come in groups of $4$, corresponding to the lower-resolution points. More precisely, rows $4k$ to $4(k+1)$ in the tensor storing the high resolution point cloud are the coordinates of the points making up the cluster that becomes row $k$ in the lower-resolution datapoint.
Since the original data is of higher dimension, we randomly subsample $2048$ points. We do this subsampling $5$ times, creating $5$ different point clouds from the same sample to act as a regularization. We adjust the epoch count accordingly when training to ensure the models are compared at the same number of iterations.
\paragraph{Objaverse-XL Preprocessing}
We utilize the Objaverse-XL dataset of meshes \cite{objaverse-xl} with captions provided by the authors of Cap-3D \cite{luo2023scalable} to create a training dataset that contains point clouds of objects of furniture. We filter the captions using a Qwen2.5 \cite{qwen2025qwen25technicalreport} language model to select objects with captions that describe a single object of furniture. We then download the meshes and sample $15000$ points from the surface of the mesh of every object to match the format of the ShapeNet dataset. The subsequent preprocessing follows the same steps as ShapeNet processing, subsampling random 8192 points and pre-computing the downsampled data with a downsampling ratio of $8$.
\paragraph{Neural Network Architecture}

We use separate PVCNN models \cite{pvcnn} for each stage, employing the same architecture in both. PVCNN combines the permutation invariance of voxel representations with the fine-grained detail of point-based networks, drawing from the design of PointNet++ \cite{pointnetplusplus}. Our model uses an embedding dimension of 64 and a voxel grid resolution of $32 \times 32 \times 32$ for the fine-grained model, and experiment with both $32 \times 32 \times 32$ and $8 \times 8 \times 8$ for the coarse-grained model. Following the standard PVCNN implementation, we apply sinusoidal timestep embeddings, a dropout rate of 0.1, and use point coordinates in place of explicit voxel features. The voxelized representation is inherently permutation-invariant, as it aggregates point-wise information into a structured volumetric form. Importantly, the computational cost of PVCNN is dominated by the convolutional operations over the voxel grid, making the voxel resolution, rather than the number of input points, the primary factor influencing runtime.

\paragraph{Hyperparameters}
The hyperparameters used for training and evaluation are listed in Table~\ref{training-hyperparams} and Table~\ref{evaluation-hyperparams}, respectively. The voxel resolution denotes the size of the voxelized representation used in the PVCNN \cite{pvcnn}. For example, a voxel resolution of 32 corresponds to a volumetric representation of shape $32 \times 32 \times 32$. The time scheduling refers to the strategy for selecting timesteps during training. We find that uniformly sampling timesteps is less effective than using a skewed distribution, which we implement via a custom scheduler denoted as sqrt scheduler $\gamma$:
\begin{equation}
    \gamma (t) = \sqrt{t}, \quad t\sim \text{Uniform}[0,1].
\end{equation}
\begin{table}[ht]
\centering
\begin{tabular}{|c|c|}
\hline
\textbf{Parameter} & \textbf{Value} \\
\hline
\hline
Optimizer & Adam \\
\hline
lr & $2e^{-4}$ \\
\hline
lr scheduling & Exponential \\
\hline
EMA & 0.9999 \\
\hline
Stage 1: Grad Clip & 0.01 \\
\hline
Time Scheduling & sqrt \\
\hline
Voxel Resolution & 32 \\
\hline
Time Boundary & 0.6 \\
\hline
Stage 1: Batch Size & 128 \\
\hline
Stage 0: Batch Size & 256 \\
\hline
Stage 1: Epochs & 300 \\
\hline
Stage 0: Epochs & 300 \\
\hline
Training Time & $\sim$ 7hrs \\
\hline
GPU & NVIDIA H100 x1 \\

\hline
\end{tabular}
\caption{Hyperparameters for training on ShapeNet.}
\label{training-hyperparams}
\end{table}

\begin{table}[ht]
\centering
\begin{tabular}{|c|c|}
\hline
\textbf{Parameter} & \textbf{Value} \\
\hline
\hline
Discretization & Euler \\
\hline
Stage 1 Time Bounds & $0-1$ \\
\hline
Stage 0 Time Bounds & $0.6-1$ \\
\hline
Stage 1: Timesteps & $1000$ \\
\hline
Stage 0: Timesteps & $400$ \\
\hline
Maximum Samples & $1000$ \\

\hline
\end{tabular}
\caption{Settings for evaluations ShapeNet.}
\label{evaluation-hyperparams}
\end{table}

\paragraph{Evaluation Metric}
We evaluate our model using 1-Nearest Neighbor Accuracy (1-NNA) under the Chamfer Distance (CD) and Earth Mover's Distance (EMD) metrics. For each evaluated class, we generate a number of samples equal to the size of the test set and compute the metrics between the generated and ground-truth test set. For the full dataset consisting of 55 categories—denoted as \textit{all}—we randomly sample 1,000 point clouds from the test set and generate 1,000 corresponding samples from the model. Metrics are then computed on these two sets. Our evaluation protocol strictly follows the procedure established in LION \citep{lion}.

\section{Additional Experimental Results}\label{append:additional-results}

\begin{table}[ht]
\centering
\caption{Ablation studies on the number of stage $K$.} 
\label{tab:3-stage}
\scalebox{0.76}{
\begin{tabular}{c|cc|cc}
\toprule & \multicolumn{2}{c|}{\textbf{Airplane}} & \multicolumn{2}{c}{\textbf{Chair}}  \\
\textbf{$K$} & CD ($\downarrow$) & EMD ($\downarrow$) & CD ($\downarrow$) & EMD ($\downarrow$) \\
\midrule
2-stage & 69.62 & 61.48 & 60.57 & 57.60 \\
3-stage & 69.40 & 59.89 & 60.78 & 58.67 \\
\bottomrule
\end{tabular}}
\end{table}
\begin{figure*}[t]
    \centering
    \includegraphics[width=0.8\textwidth]{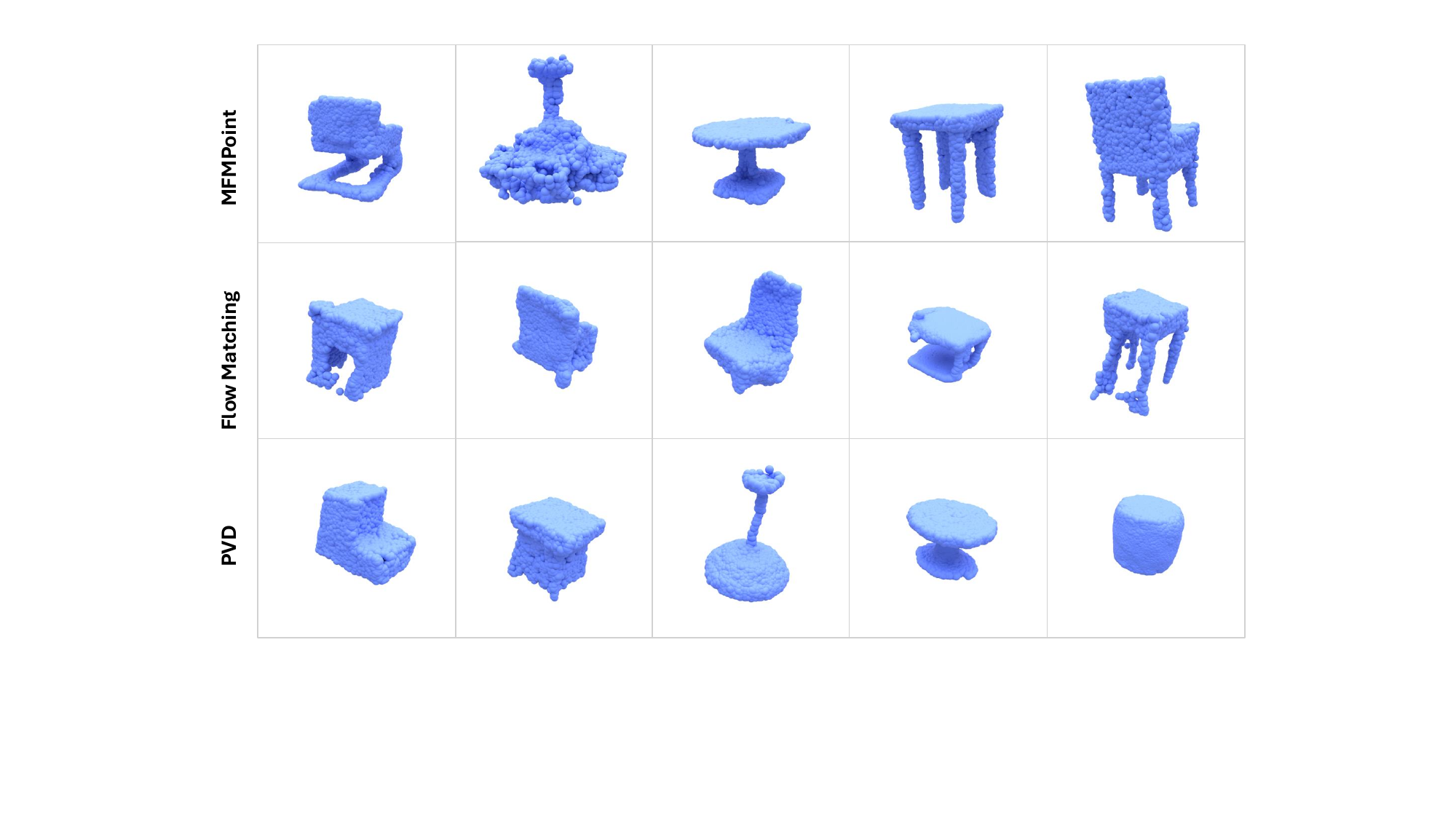}
    \vspace{-50pt}
    \caption{Side-by-side qualitative comparison between MFM-point, FM and PVD.}
    \label{fig:furniture_uncurated}
\end{figure*}
\paragraph{Ablation Studies on the number of stages $K$}
We further investigate the impact of the number of stages $K$ in our multi-scale framework through ablation experiments. We evaluate a 3-staged model with transition time set to $t^1_s=0.3, t^0_s = 0.6$. We compare its performance against our default 2-stage model on high-resolution (8192 points) Airplane and Chair datasets. As shown in Table \ref{tab:3-stage}, the 3-stage model achieves a performance comparable to the 2-stage model, indicating that increasing the number of stages does not necessarily lead to significant gains in this case.


\section{Additional Qualitative Results}\label{appen:qualitative}
In this section, we present uncurated samples generated by our MFM-point.


\clearpage

\begin{figure*}[t]
    \centering
    \includegraphics[width=\textwidth]{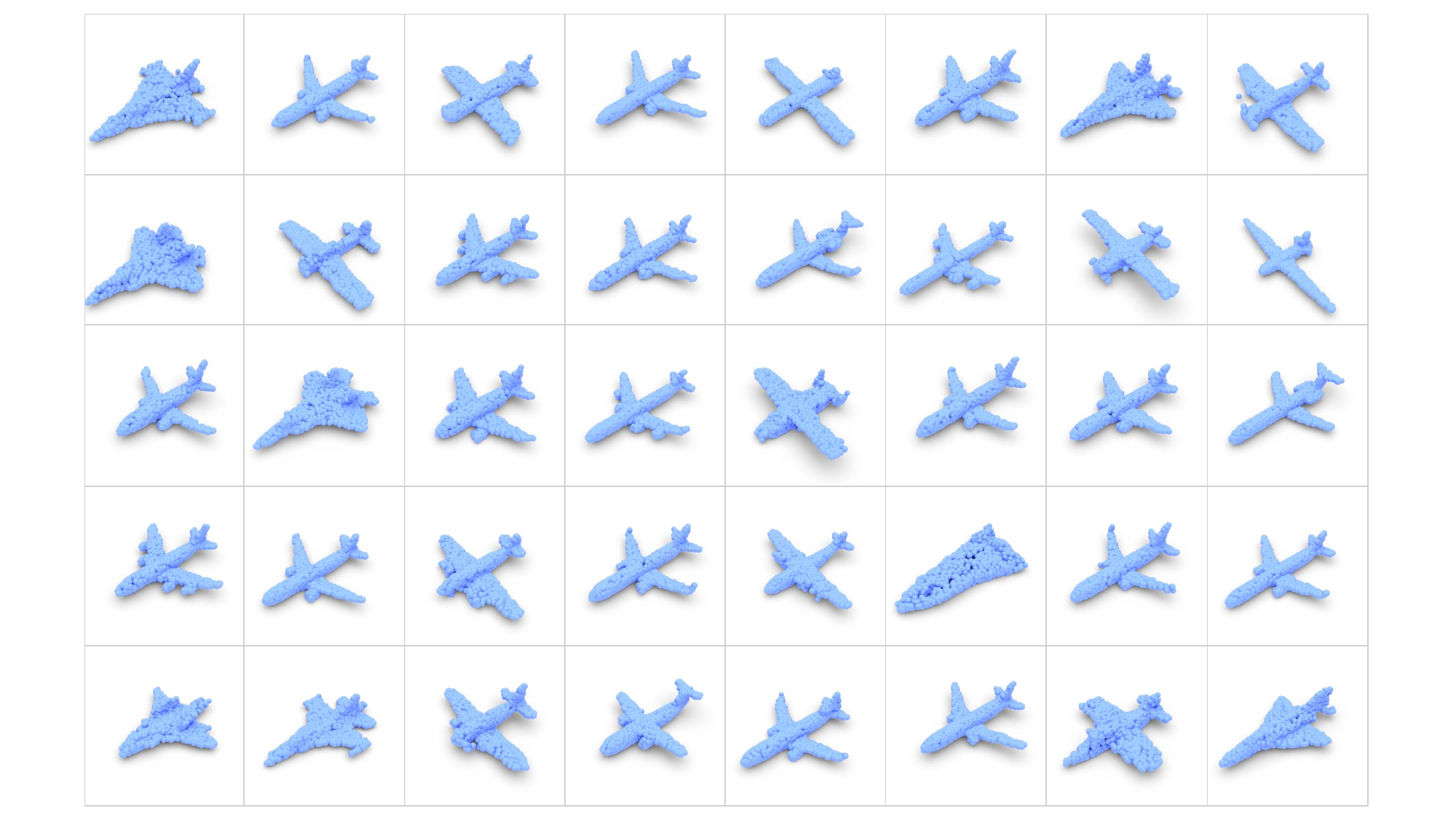}
    \caption{Generated samples from our 2-stage model in the single-category setting for the airplane class.}
\end{figure*}

\begin{figure*}[t]
    \centering
    \includegraphics[width=\textwidth]{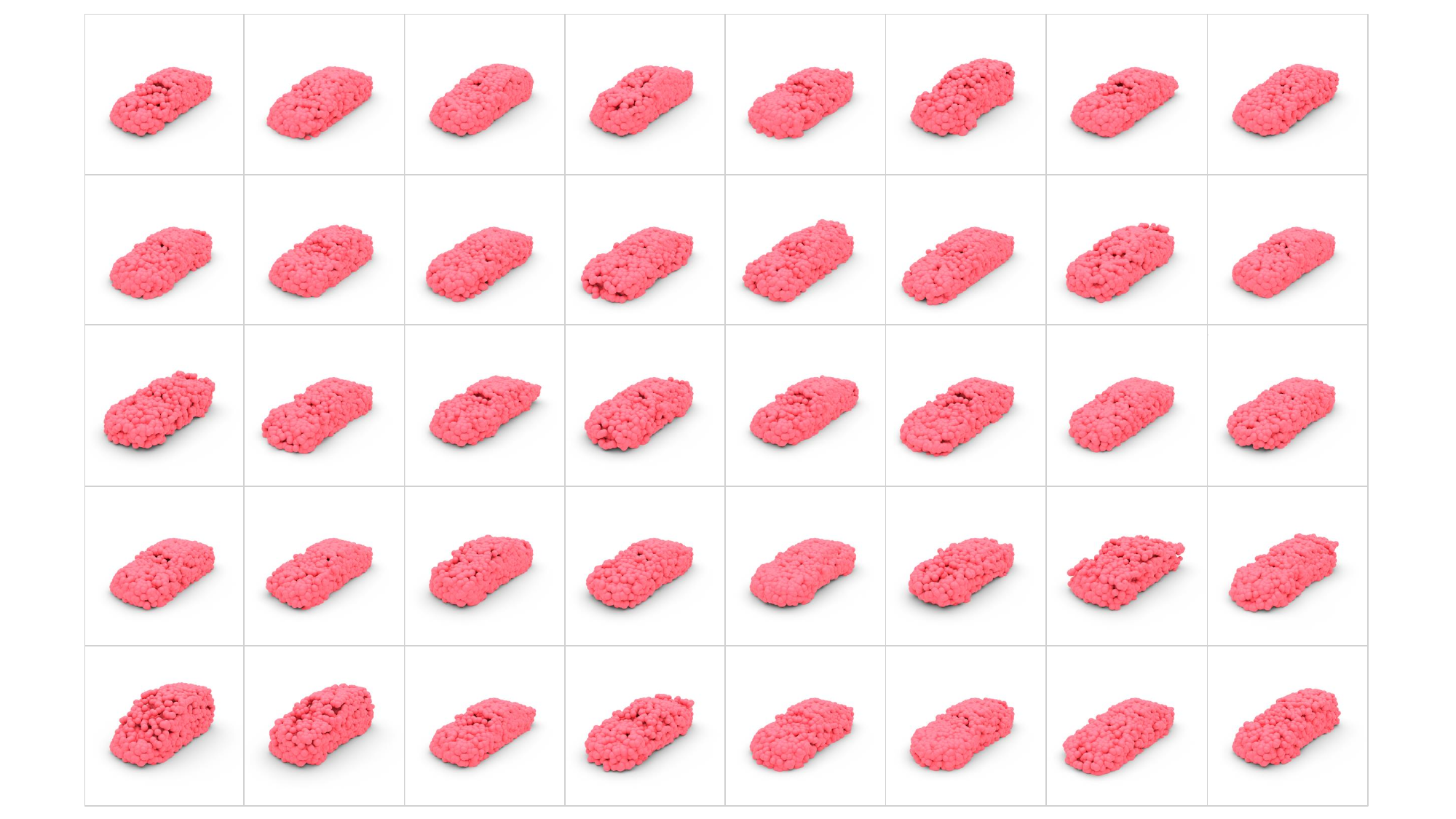}
    \caption{Generated samples from our 2-stage model in the single-category setting for the car class.}
\end{figure*}

\begin{figure*}[t]
    \centering
    \includegraphics[width=\textwidth]{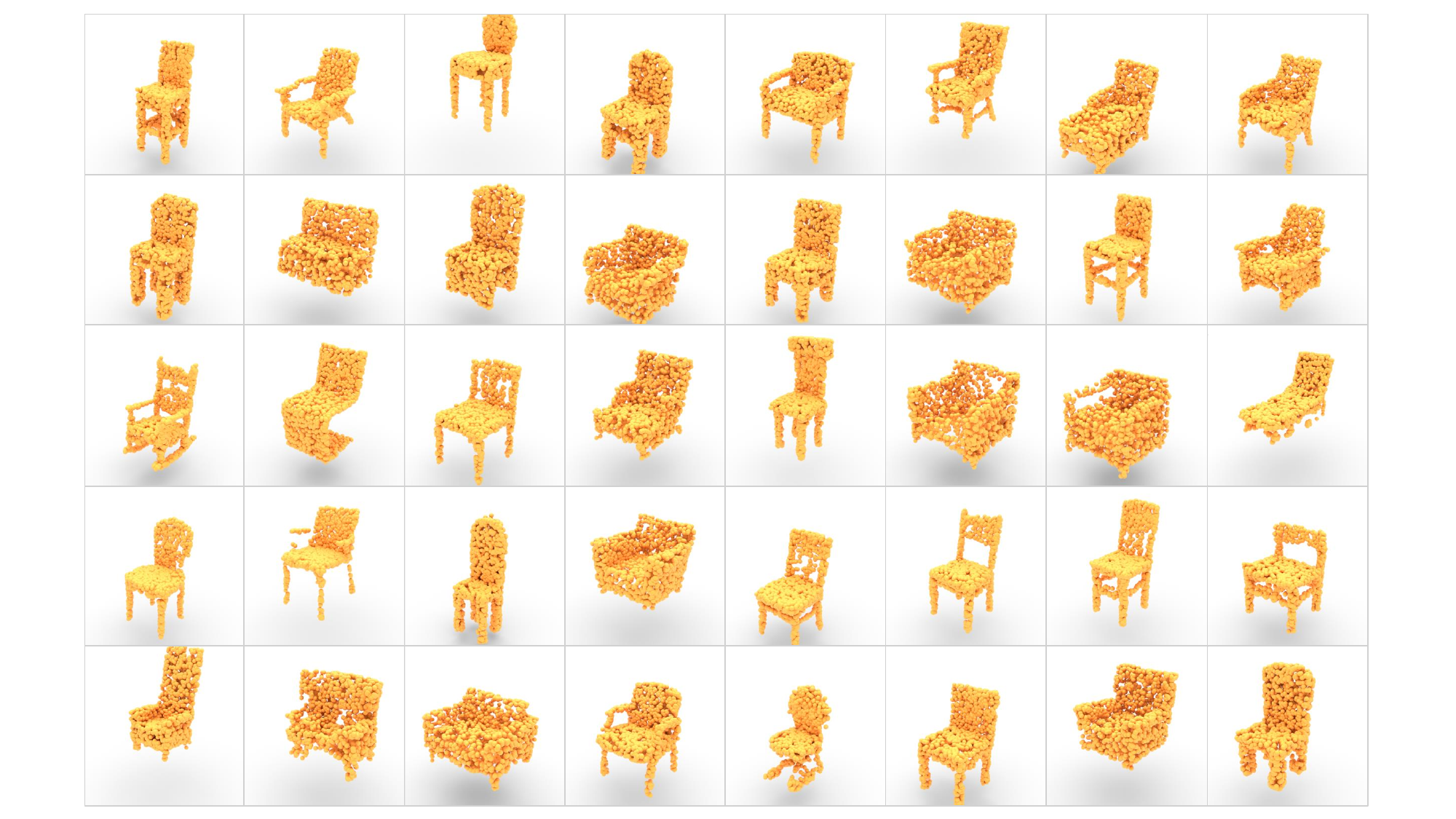}
    \caption{Generated samples from our 2-stage model in the single-category setting for the chair class.}
\end{figure*}

\begin{figure*}[t]
    \centering
    \includegraphics[width=\textwidth]{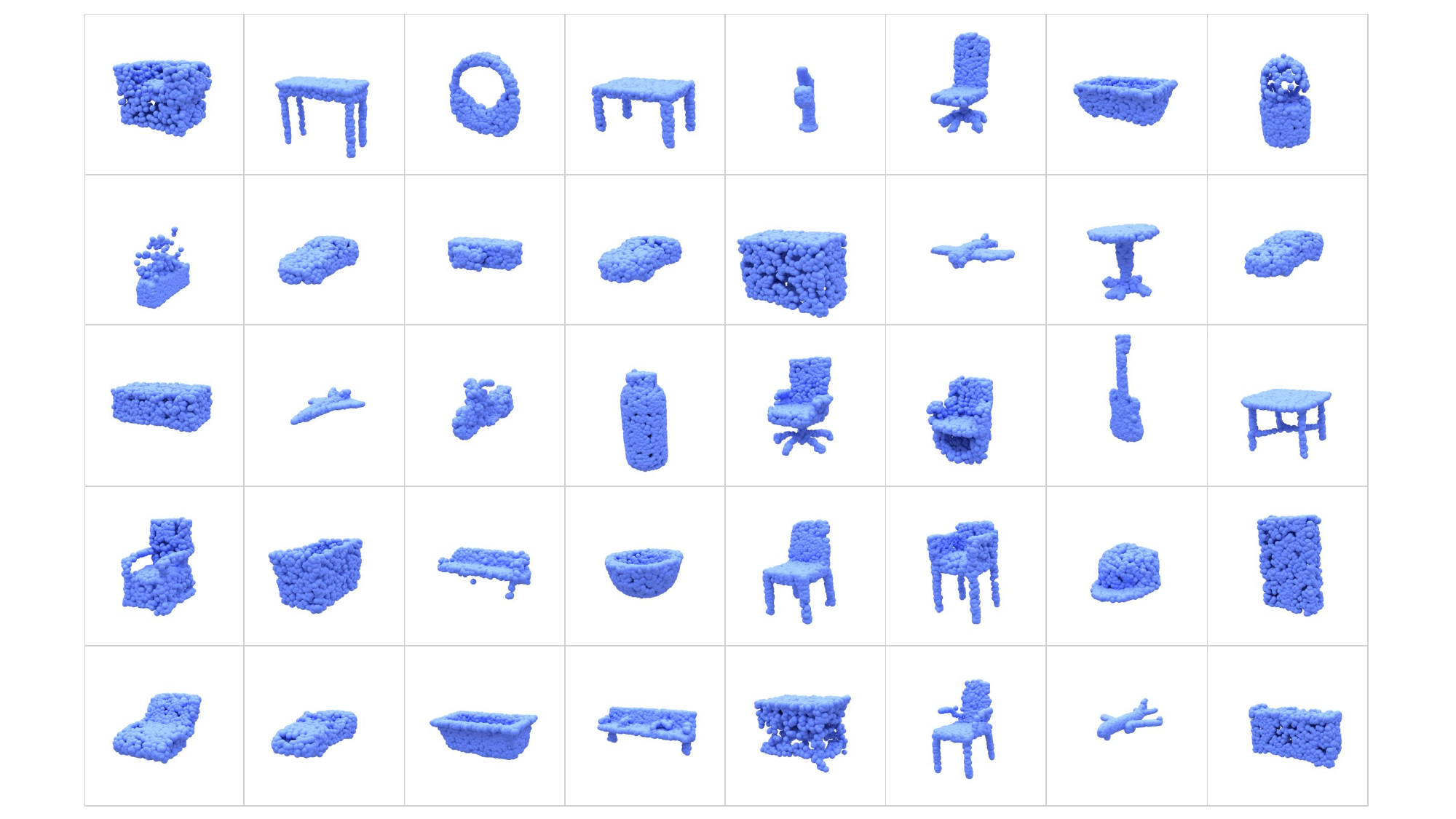}
    \caption{Generated samples from our 2-stage model in the unconditional multi-category setting.}
\end{figure*}

\begin{figure*}[t]
    \centering
    \includegraphics[width=\textwidth]{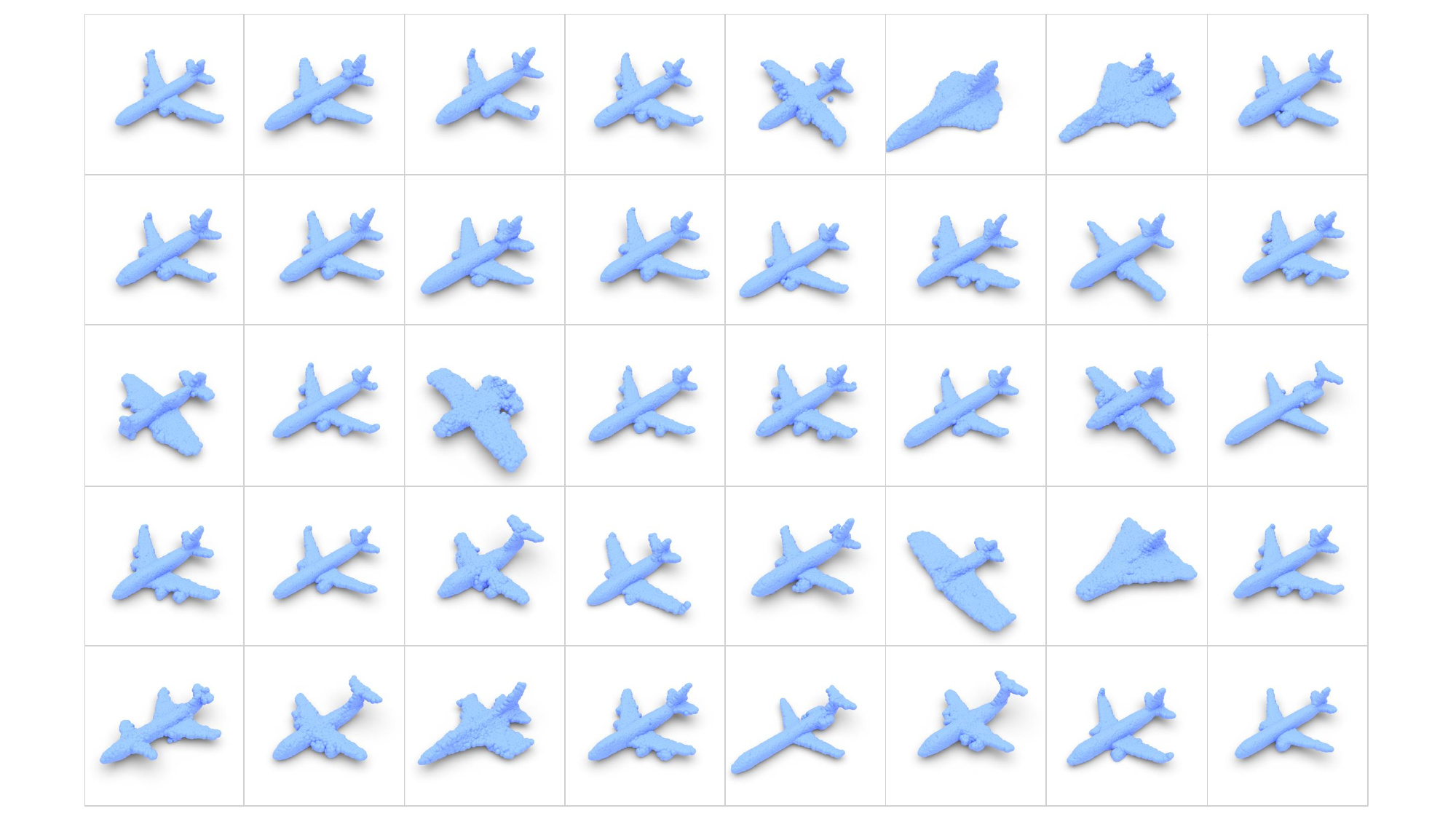}
    \caption{Generated samples from our 2-stage model in the single-category setting for the airplane class for \textbf{8192 point} resolution.}
\end{figure*}

\begin{figure*}[t]
    \centering
    \includegraphics[width=\textwidth]{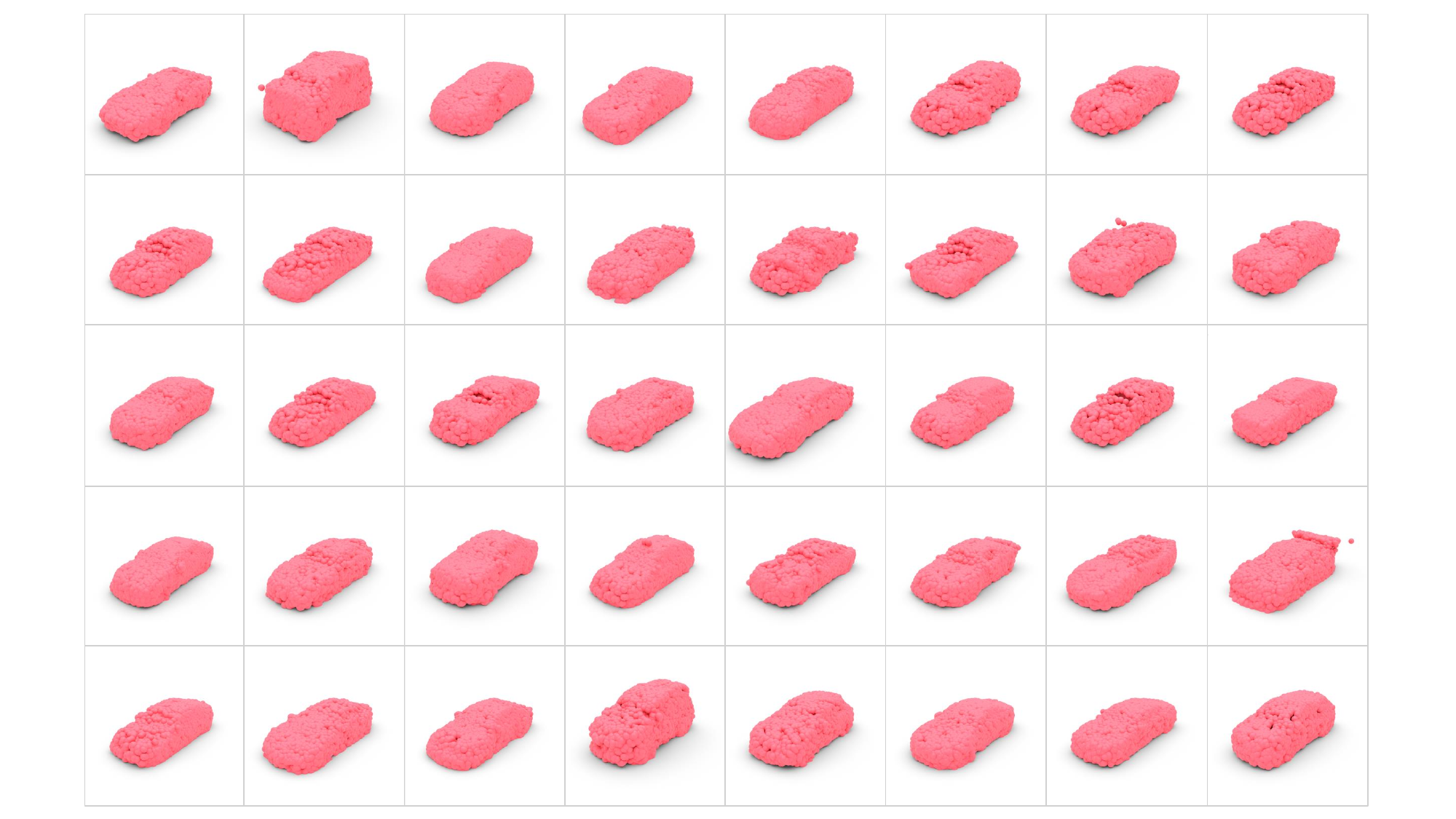}
    \caption{Generated samples from our 2-stage model in the single-category setting for the car class for \textbf{8192 point} resolution.}
    \label{fig:car_8192_uncurated}
\end{figure*}

\begin{figure*}[t]
    \centering
    \includegraphics[width=\textwidth]{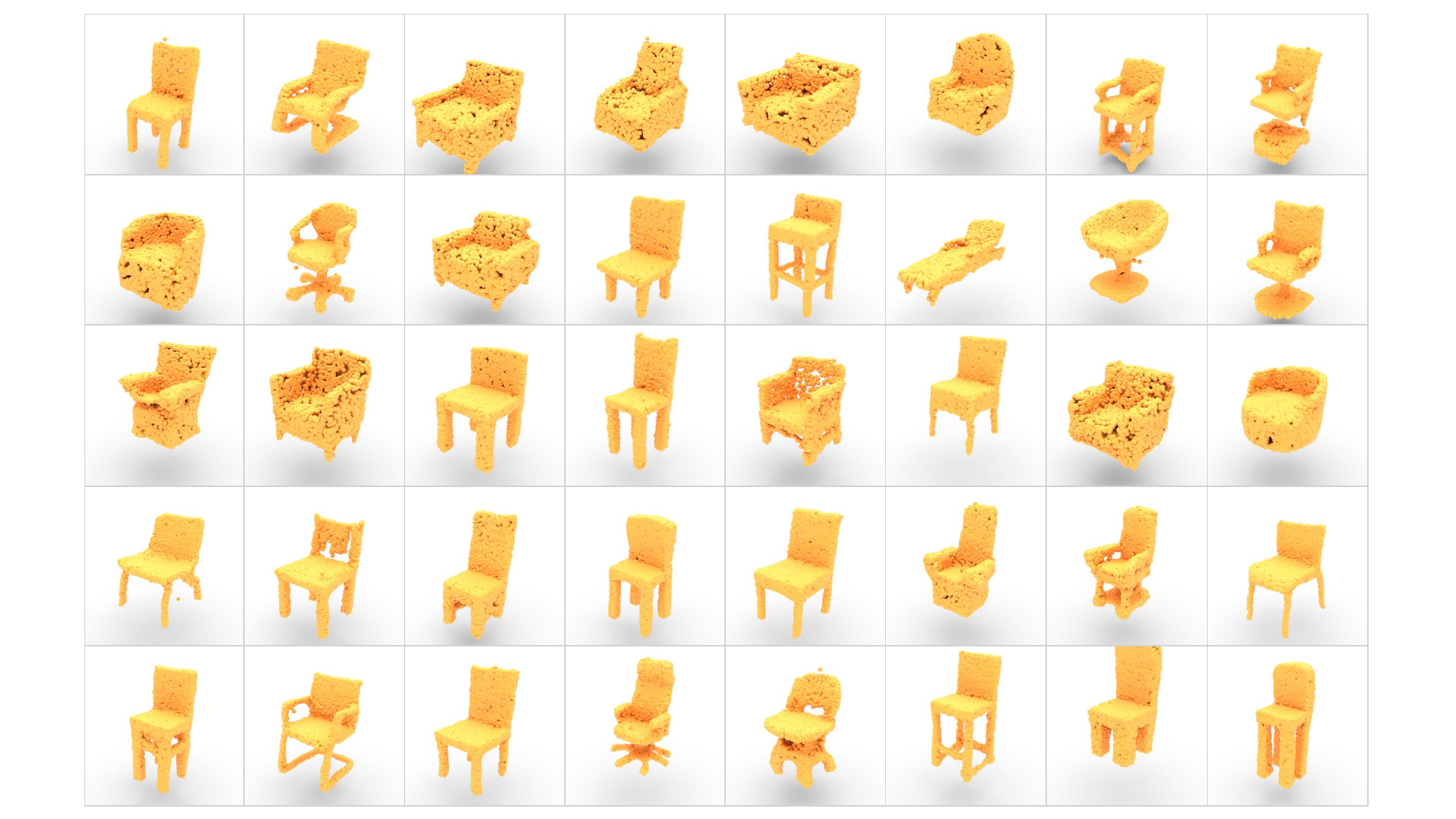}
    \caption{Generated samples from our 2-stage model in the single-category setting for the chair class for \textbf{8192 point} resolution.}
    \label{fig:chair_8192_uncurated}
\end{figure*}

\begin{figure*}[t]
    \centering
    \includegraphics[width=\textwidth]{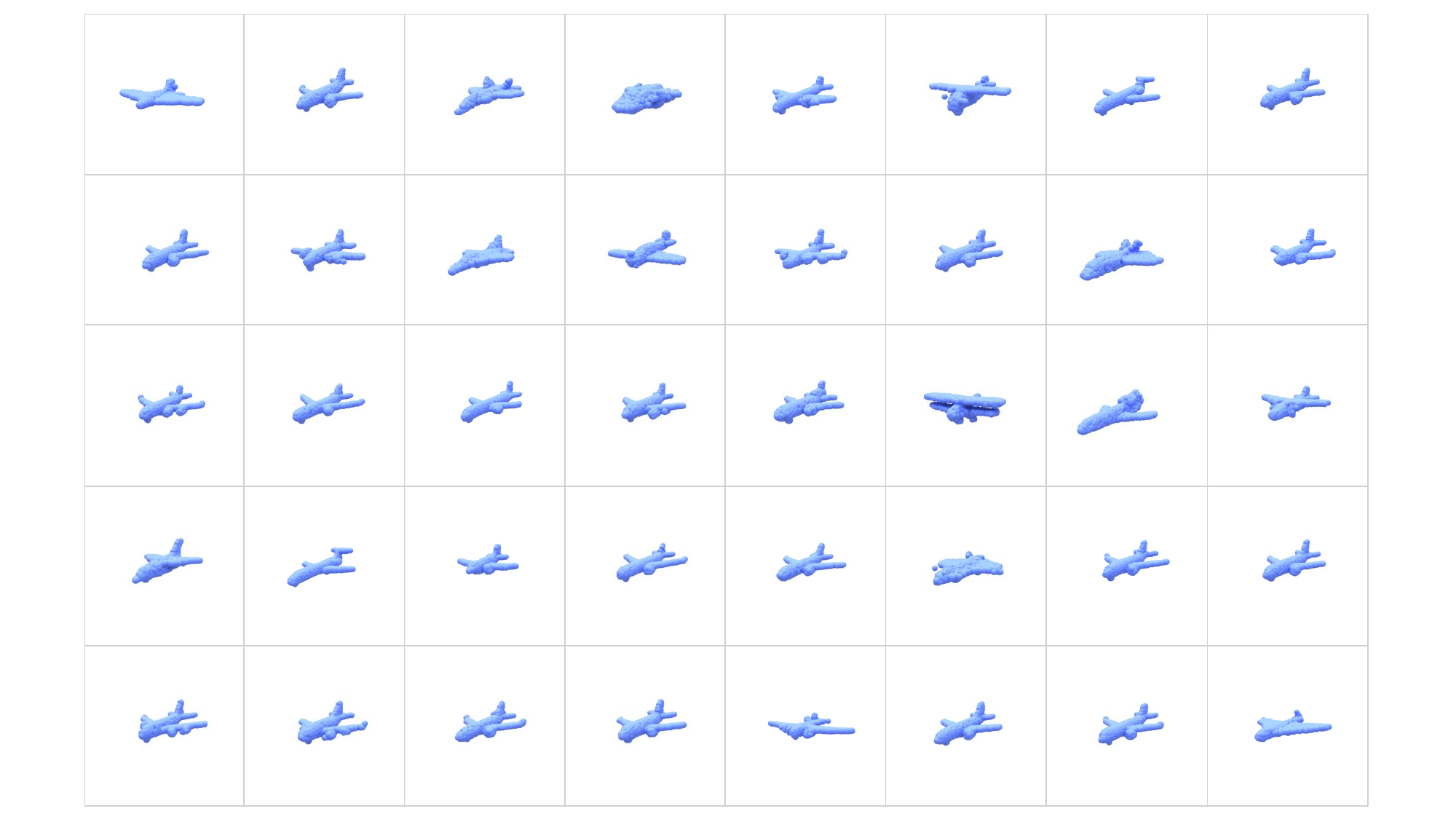}
    \caption{Generated samples from our 2-stage model in the \textbf{conditional} single-category setting for the airplane class.}
\end{figure*}

\begin{figure*}[t]
    \centering
    \includegraphics[width=\textwidth]{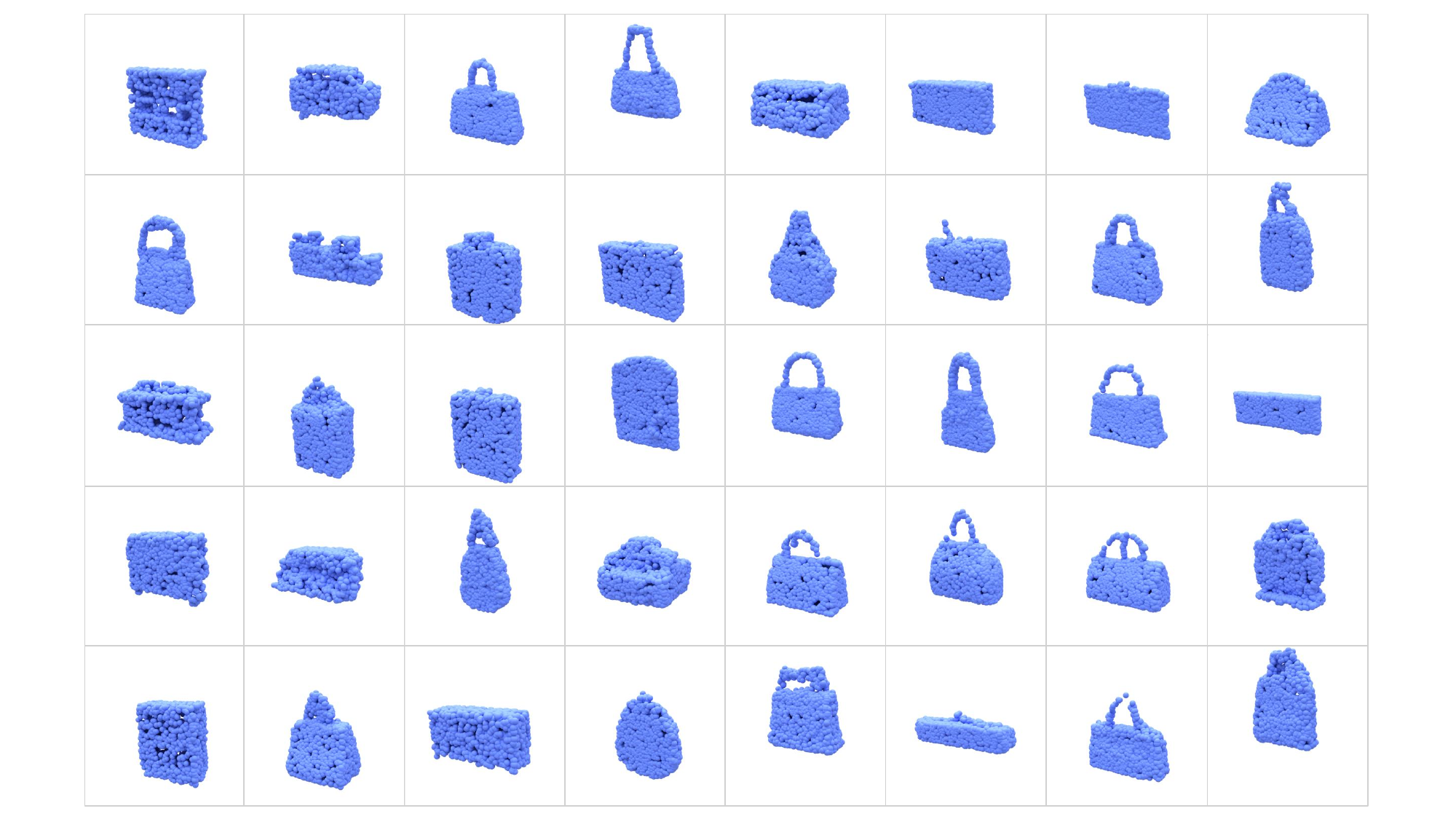}
    \caption{Generated samples from our 2-stage model in the \textbf{conditional} single-category setting for the bag class.}
\end{figure*}

\begin{figure*}[t]
    \centering
    \includegraphics[width=\textwidth]{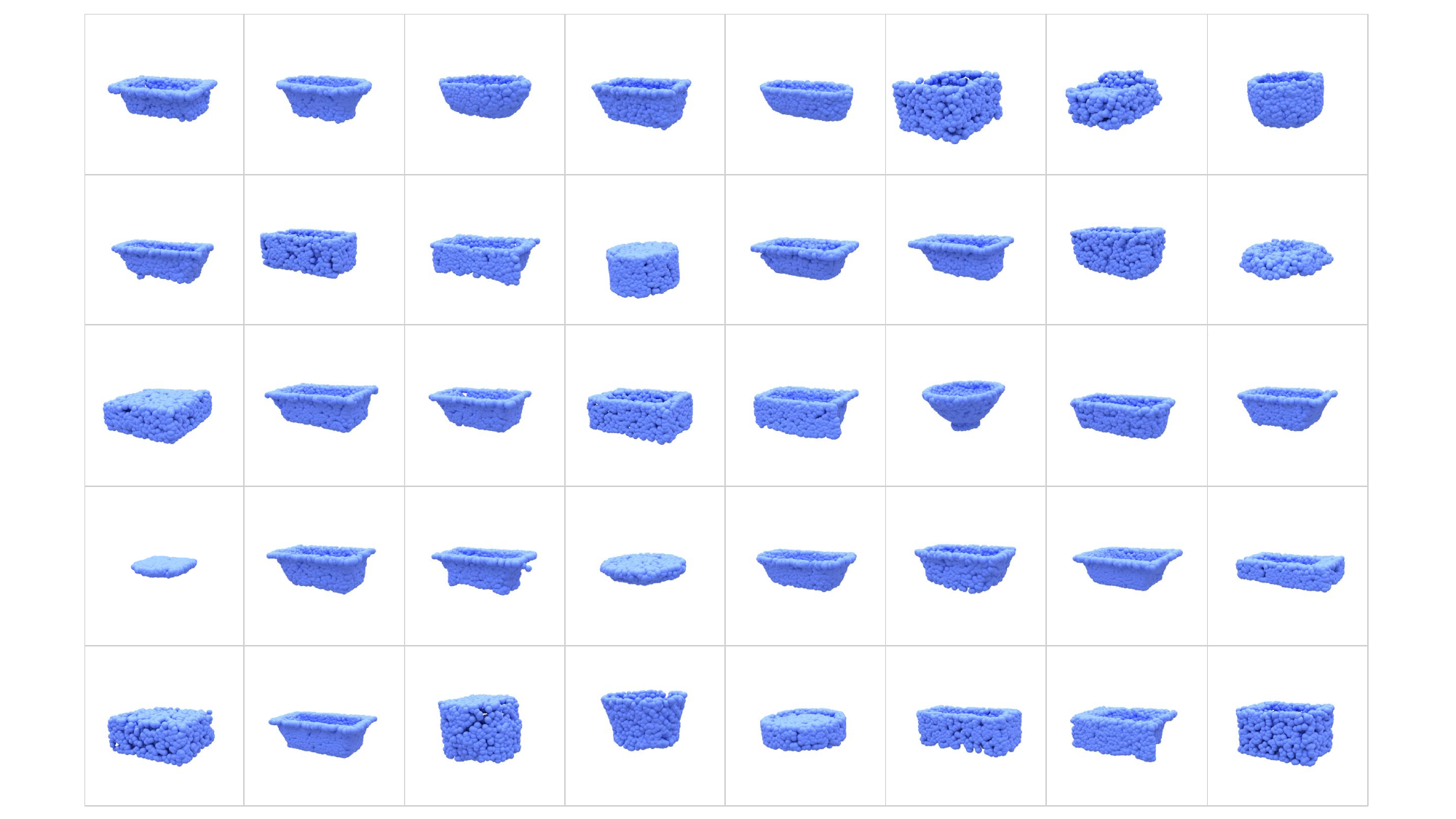}
    \caption{Generated samples from our 2-stage model in the \textbf{conditional} single-category setting for the bathtub class.}
\end{figure*}

\begin{figure*}[t]
    \centering
    \includegraphics[width=\textwidth]{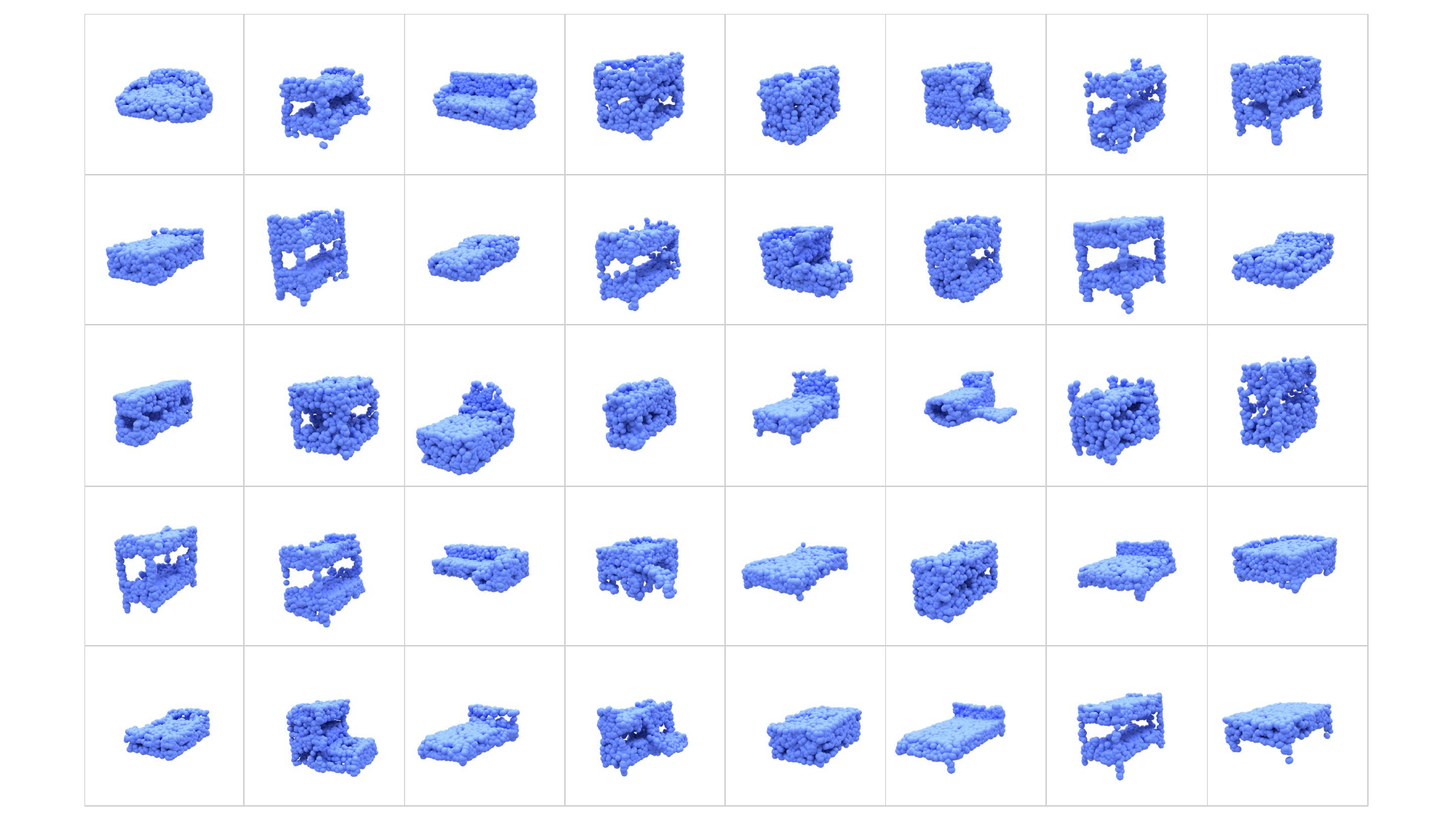}
    \caption{Generated samples from our 2-stage model in the \textbf{conditional} single-category setting for the bed class.}
\end{figure*}

\begin{figure*}[t]
    \centering
    \includegraphics[width=\textwidth]{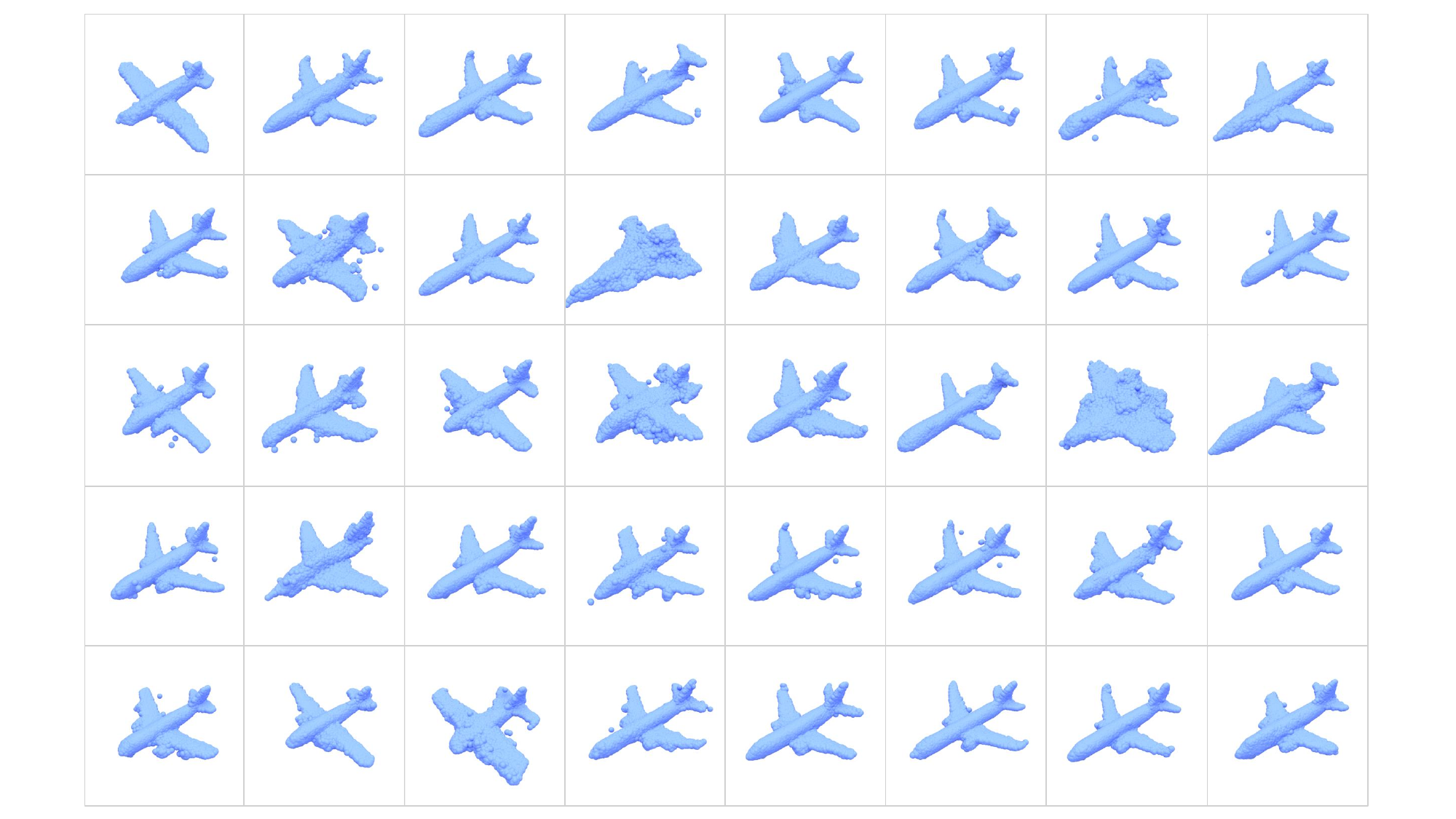}
    \caption{Generated samples from our 2-stage model in the single-category setting for the airplane class for \textbf{15000 point} resolution.}
    \label{fig:airplane_15k_uncurated}
\end{figure*}

\end{document}